\documentclass[10pt,number,sort&compress]{elsarticle}
\usepackage{times} 
\usepackage{courier} 
\usepackage[margin=1.25in]{geometry}
\usepackage{amsmath, amssymb}
\usepackage{graphicx}
\usepackage[normalem]{ulem}
\usepackage{color}
\usepackage{xcolor}
\usepackage[utf8]{inputenc}
\usepackage{qcircuit} 
\usepackage{xcolor}
\usepackage[font=footnotesize,labelfont=bf]{caption}
\usepackage[font=footnotesize,labelfont=bf]{subcaption}
\usepackage{caption}  
\usepackage{subcaption} 
\usepackage{siunitx}
\usepackage{hyperref}
\usepackage{algorithm}
\usepackage{movie15}
\usepackage{setspace}
\usepackage{enumitem}
\usepackage{booktabs}
\usepackage{comment}
\usepackage{threeparttable}
\DeclareUnicodeCharacter{2217}{\ensuremath{*}}

\newcounter{appsection}
\renewcommand{\theappsection}{A\arabic{appsection}}
\newcommand{\appsec}[1]{%
  \refstepcounter{appsection}%
  \section*{Appendix \theappsection. #1}%
}

\newcounter{supsection}
\renewcommand{\thesupsection}{\Alph{supsection}}
\newcommand{\supsec}[1]{%
  \refstepcounter{supsection}%
  \section*{Supplemental Material-\thesupsection. #1}%
}

\begin{document}
\begin{frontmatter}
\title{On Surrogate Modeling of Static Response of AM Short-Fiber Thermoplastics Using Graph Neural Networks}

\author[add1]{Pharindra Pathak\corref{correspondingauthor}}
\author[add2]{Vipin Kumar}
\author[add3]{Trenton M. Ricks}
\author[add4]{Suhasini Gururaja}
\author[add5]{Siddhartha Srivastava}
\address[add1]{Graduate Student, Department of Aerospace Engineering, Auburn University, Auburn, AL}
\address[add2]{Sr. R\&D Staff Member, Manufacturing Science Division, ORNL, Knoxville, TN}
\address[add3]{Research Aerospace Engineer, Multiscale and Multiphysics Modeling Branch, NASA Glenn Research Center, Cleveland, OH}
\address[add4]{Professor, Department of Aerospace Engineering, Auburn University, Auburn, AL}
\address[add5]{Assistant Professor, Department of Aerospace Engineering, Auburn University, Auburn, AL}
\cortext[correspondingauthor]{Corresponding author: Pharindra Pathak, Graduate Student, Department of Aerospace Engineering, Auburn University, Email: pzp0057@auburn.edu}

\begin{abstract}
Short-fiber thermoplastic (SFT) composites are increasingly employed in lightweight aerospace and automotive structures owing to their favorable strength-to-weight ratio, high production rates, and recyclability. Unlike continuous-fiber systems, the mechanical response of SFTs is governed by complex mesoscale interactions among fiber orientation, spatial clustering, and manufacturing-induced porosity. These microstructural features exhibit significant spatial variability in additively manufactured components and strongly influence stiffness, damage initiation, and nonlinear deformation. Although mesoscale finite element (FE) models can explicitly resolve such heterogeneity, their application to realistic three-dimensional microstructures containing thousands of fibers remains computationally intractable.

In this study, a data-driven surrogate modeling framework is proposed to efficiently predict the mechanical behavior of additively manufactured, compression-molded (AM-CM) SFTs. Three-dimensional microstructures were reconstructed from high-resolution micro-computed tomography ($\mu$-CT) data and discretized into localized Voronoi-based cells representing statistically distinct fiber-interaction neighborhoods. Each cell was homogenized via nonlinear FE simulations incorporating matrix damage. The resulting stress-strain responses were used to train a hybrid Graph Neural Network-Long Short-Term Memory (GNN-LSTM) architecture capable of simultaneously encoding microstructural topology and capturing nonlinear, history-dependent mechanical evolution.

The trained surrogate accurately predicts the stiffness and stress-strain behavior of unseen microstructures, with a coefficient of determination $R^2 \approx 0.98$ relative to high-fidelity FE simulations, while achieving more than two-order-of-magnitude reduction in computational cost. Coupling the framework with experimentally calibrated static damage laws demonstrates that fiber orientation, clustering, and porosity collectively govern the local effective stiffness matrix. The proposed approach provides a physics-informed, data-efficient pathway to identify mechanically weak microstructural cells and accelerate digital-twin development for SFT components.
\end{abstract}


\end{frontmatter}
\begin{doublespace}

\section{Introduction}
Short-fiber thermoplastics (SFTs) are increasingly deployed in lightweight aerospace and automotive structures for their high production rates, recyclability, and competitive mechanical properties relative to traditional thermoset-based composites \cite{billah_2024, Humberto_2025, pathak_porosity_2025}. Recently, hybrid manufacturing that combines additive manufacturing (AM) with compression molding (CM) has emerged as a promising processing strategy. This AM-CM approach mitigates residual porosity, enhances inter-bead bonding, and improves consolidation in pellet-extrusion systems. During AM, intense shear flow within the extrusion nozzle induces preferential fiber alignment along the deposition path. The subsequent compression molding step, performed above the matrix melt temperature, reduces global porosity by approximately 50\texorpdfstring{\%} and strengthens fiber-matrix interfacial bonding \cite{kumar_high-performance_2021,rezaei_development_2008,donal_1997,gandhi_method_2016,tekinalp_highly_2014,parandoush_review_2017,advani_3_2012}. 

Despite these processing advantages, AM-CM SFT components still exhibit pronounced microstructural heterogeneity \cite{donal_1997,arif_multiscale_2014, Almeida, stamopoulos_evaluation_2016}. Unlike continuous-fiber laminates with layered construction, SFT microstructures exhibit spatial variability in fiber orientation, fiber clustering, and porosity due to complex flow kinematics during deposition and molding \cite{yu_3d_2020, donal_1997}. From a micromechanics perspective, these variations directly alter local load-transfer efficiency. In SFTs, stress redistributes through matrix shear (shear lag), making axial stress transfer highly sensitive to fiber proximity, alignment, and termination. Regions of fiber clustering and fiber-end neighborhoods introduce stress concentrations, amplify matrix shear stresses, and act as preferred sites for damage initiation under both monotonic and cyclic loading \cite{Horst_1996,fu_post-mortem_2005, Tamboura_2020, Mortazavian_2015}.

In our previous work on AM-CM chopped carbon-fiber-reinforced acrylonitrile butadiene styrene (CF-ABS) SFTs, we confirmed a strong correlation between microstructural topology and mechanical performance \cite{pathak_porosity_2025}. High-resolution micro-computed tomography ($\mu$-CT) and interrupted fatigue testing revealed that porosity nucleates and grows preferentially near fiber termini and within fiber clusters during cyclic loading. Fiber ends disrupt matrix continuity and create stress concentrations, while dense fiber clusters restrict matrix flow, forming mechanically weak zones. As the number of fatigue cycles increases, secondary microstructural evolution---including fiber-matrix debonding, fiber rotation, and local rearrangement---modifies both the fiber length distribution (FLD) and the fiber orientation distribution (FOD). Quasi-static tensile tests performed after fatigue interruption reveal reduced stiffness and an earlier onset of nonlinearity relative to the pristine state, confirming that the mechanical response of AM-CM SFTs is primarily governed by the local topology of fiber interactions.

Mean-field homogenization approaches, including Mori--Tanaka and related inclusion-based models, offer computational efficiency by spatially averaging reinforcement contributions within an equivalent homogeneous matrix \cite{Abhilash_2024, Pothnis_2019, Kammoun_2015, Breuer_2021, Hao_2022}. Because these formulations rely on spatial averaging, they cannot explicitly represent fiber-fiber interactions, resin-rich pockets, or localized damage hotspots that are directly observable in $\mu$-CT-resolved SFT microstructures \cite{muller_homogenization_2015, kaiser_extended_2012, li_mean-field_2025}. While such models can provide reasonable estimates of global elastic properties, they remain fundamentally limited in capturing microstructure-driven nonlinear response and progressive stiffness degradation \cite{arif_multiscale_2014, uhlig_meso-scaled_2016, estefani_numerical_2024}.

At the other end of the spectrum, fully resolved three-dimensional finite element (FE) models based on $\mu$-CT data can capture local stress concentrations and load-transfer pathways. However, the complexity of realistic SFT microstructures---which feature hundreds to thousands of fibers---necessitates extremely large meshes, resulting in prohibitive computational costs, particularly when accounting for nonlinear plasticity and damage. This restricts their use for parametric studies, uncertainty quantification, and multiscale modeling.

Surrogate modeling offers a middle ground between fine-scale and coarse-scale approaches by decomposing heterogeneous microstructures into smaller sub-domains, hereafter referred to as cells, each representing a statistically homogeneous local fiber-interaction neighborhood that retains key micromechanical features while reducing computational effort~\cite{wu_micro-mechanics_2021, schraa_characterisation_2025, mentges_micromechanical_2023, hamza_physics-constrained_2025, kobler_fiber_2018,eyri_modeling_2025, Senthilnathan_2024, chinesta_short_2011, dai_graph_2021, summerscales_voronoi_2001, gulmez_quantification_2023}. However, when each cell still requires nonlinear FE homogenization, a surrogate does not fully close the computational gap. Scalability remains limited, particularly as finer spatial resolution is required to capture microstructural variability.

Whereas the surrogate strategies discussed above reduce cost through geometric decomposition while still relying on FE evaluation of each cell, data-driven surrogate models based on machine learning aim to replace the FE solver itself with a learned mapping from microstructure to response~\cite{cassola_machine_2022,sharma_review_2025,ferdousi_deep_2025,kibrete_ai_2023,chen_machine_2019}. Image-based convolutional neural networks and graph-based learning methods have demonstrated promising capabilities for predicting effective properties or stress fields in simplified composite systems~\cite{ge_numerical_2024, yacouti_integrated_2025, dai_graph_2021, vitulyova_hybrid_2025}. However, many existing frameworks rely on synthetic microstructures and do not explicitly represent the physical topology of fiber-fiber interactions present in real $\mu$-CT data. Furthermore, most existing frameworks are limited to state-based property prediction~\cite{Breuer_2021, zhao_high-generalizability_2023, Senthilnathan_2024}, in which the response is assumed to depend only on the current configuration, and do not address history-dependent behavior where the stress response is governed by the entire deformation history through evolving internal variables~\cite{gunst_leveraging_2025, chen_deep_2021, jiang_design_2022, zhao_high-generalizability_2023, Mozaffar_2019, Ghaboussi_1991, Burigede_2023}.

These limitations reveal a critical modeling gap: predictive frameworks are needed that (i) preserve experimentally observed microstructural topology, including fiber clustering, fiber-end effects, and resin-rich regions; (ii) capture history-dependent nonlinear damage evolution under quasi-static loading; and (iii) remain computationally tractable for coupon-level analysis.

\begin{figure}[htb!]
    \centering
    \includegraphics[width=1\linewidth]{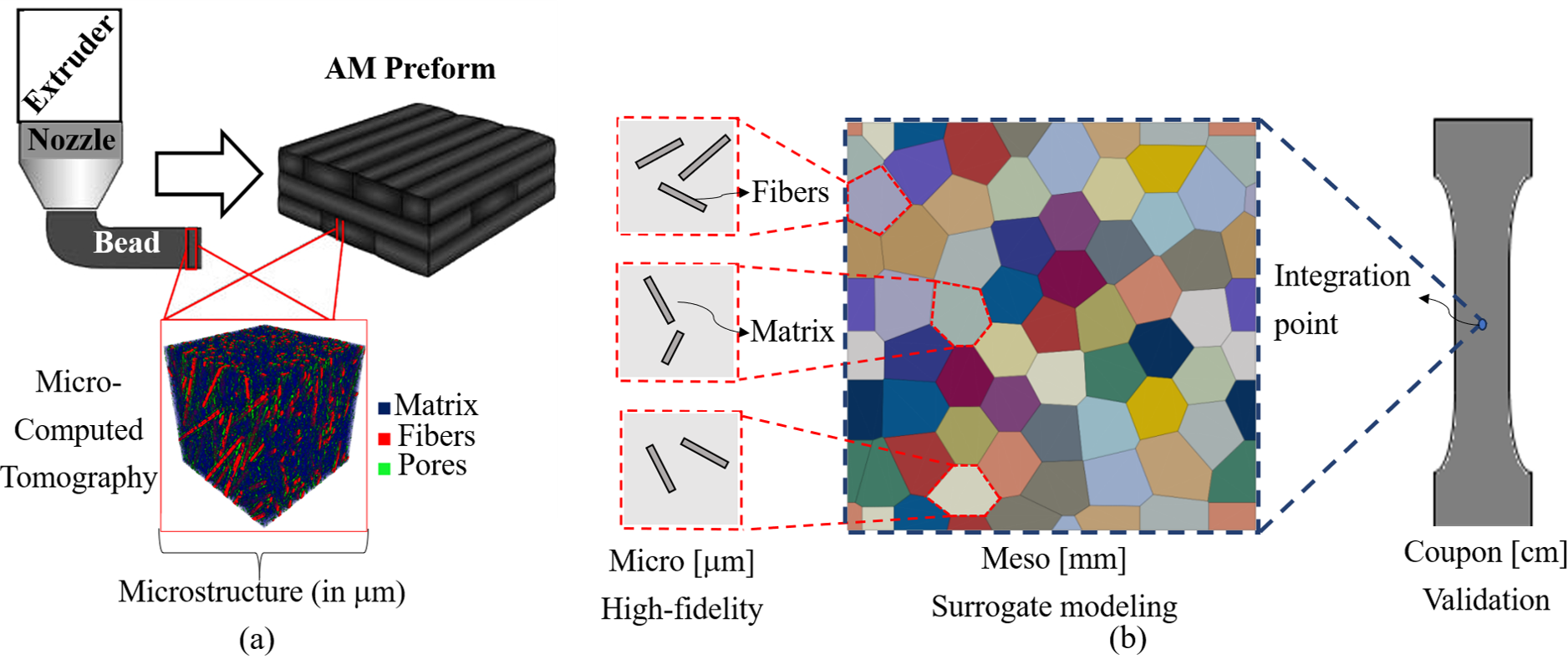}
    \caption{Overview of the multiscale modeling workflow. (a) Pellet-extrusion AM, followed by compression molding (not shown), produces a CF-ABS preform. High-resolution $\mu$-CT imaging reconstructs the 3D microstructure at the micrometer scale, revealing the matrix, fiber, and pore phases. (b) The reconstructed microstructure is partitioned into mesoscale Voronoi cells using surrogate modeling to allow high-fidelity coupon-scale validation. The workflow spans three length scales: microscale [$\mu$m] for high-fidelity FE homogenization, mesoscale [mm] for surrogate-based cell assembly, and coupon [cm] for experimental validation (adapted from \cite{pathak_porosity_2025}). \\
    \textbf{Note:} The color map is used for visual differentiation of cells and does not represent fiber count or local fiber concentration.}
    \label{fig:1}
\end{figure}

In this work, we develop a data-driven multiscale surrogate modeling framework that operates on $\mu$-CT-resolved microstructural topology, characterized by fiber geometry and their connectivity through proximity-dependent load-transfer mechanisms. The framework learns the mapping from these fiber statistics to the nonlinear, history-dependent stress--strain response while enforcing physical consistency and enabling computationally efficient predictions. Figure~\ref{fig:1} depicts the multiscale modeling workflow for AM-SFTs. The AM-CM process produces a bead-scale aligned preform characterized via high-resolution $\mu$-CT imaging. The reconstructed microstructure is partitioned into mesoscale Voronoi cells that tessellate the domain into subdomains with statistically homogeneous fiber configurations, capturing local variations in clustering and mutual proximity. Nonlinear FE homogenization is first performed to generate topology-dependent constitutive data for each cell. A hybrid Graph Neural Network--Long Short-Term Memory (GNN-LSTM) surrogate model is then trained to approximate the homogenization operator: fibers are treated as graph nodes, and load-transfer pathways as weighted edges, thereby preserving the fiber-to-fiber interactions, while the LSTM captures path-dependent plasticity and damage evolution in the matrix. Individual Voronoi cell responses are subsequently assembled to predict coupon-scale mechanical behavior while retaining proximity-driven load-transfer mechanisms.

The primary contribution of this work is the formulation of a topology-preserving surrogate model that directly links $\mu$-CT-resolved fiber configurations to nonlinear history-dependent constitutive response. Unlike conventional approaches that rely on homogenized statistical descriptors, the proposed topology-preserving framework explicitly retains local interaction mechanisms governing load transfer and damage evolution. By replacing high-fidelity nonlinear FE homogenization with a trained GNN–LSTM surrogate, the framework significantly reduces computational cost while maintaining the micromechanical fidelity required to predict stiffness degradation in AM-CM SFT systems.

\section{Materials and Methods}
\label{sec:manufacturing}
\subsection{Additive manufacturing-compression molding process}
CF-ABS panels containing 20 wt.\% chopped carbon fibers were fabricated using a two-step process developed at Oak Ridge National Laboratory, combining pellet-extrusion additive manufacturing (AM) with subsequent compression molding (CM), as described in prior studies \cite{Pathak_Comp, pathak_porosity_2025, Pathak_ffems_2026} (see Figure~\ref{fig:1}a). Unlike conventional filament-based fused deposition modeling (FDM), this approach utilizes a large-format, pellet-fed material extrusion system. During AM, a 20 wt.\% CF-ABS feedstock, comprising ABS pellets reinforced with chopped high-tenacity carbon fibers (Kaltex K20-HTU), was deposited using a KUKA robotic extrusion system.

The extrusion was performed at a screw speed of 1200 rpm, a nozzle traverse speed of 280 mm/s, and a nozzle-substrate gap of 9.88 mm, producing single-bead layers with a height of approximately 6--7 mm. The extrusion system operated at a nozzle temperature of 523 K, with four internal heating zones progressively increasing from 363 K (Zone 1) to 498 K (Zone 2), 523 K (Zone 3), and 523 K at the nozzle. This processing condition induces strong shear flow within the nozzle, promoting preferential fiber alignment along the deposition direction and resulting in a transversely anisotropic microstructure at the bead scale. Fully dense (100\% infill) additive preforms were produced with dimensions of approximately 450 mm × 300 mm × 6.5 mm. Subsequently, compression molding was performed above the ABS melt temperature (523 K) for four minutes under a controlled pressure of a 250-ton hot press (Carver Press, Model $\#$3895.4NE10000). This step enhanced inter-bead fusion, reduced global porosity, and improved fiber-matrix interfacial bonding, yielding consolidated panels with final dimensions of approximately 450 mm × 300 mm × 3.5 mm \cite{Pathak_ffems_2026, pathak_porosity_2025, Pathak_Comp}.

Despite nominally identical processing parameters, the AM-CM process inherently generates spatially variable microstructures \cite{pathak_porosity_2025, billah_2024, Vipin_2025}. Variations in local flow kinematics, thermal gradients, and consolidation pressure lead to non-uniform fiber clustering and residual porosity distributions. While the fabricated panels meet macroscopic quality metrics, localized regions may exhibit high fiber clustering, which can serve as sites of high stress concentration. From a micromechanics perspective, these features can promote early damage initiation \cite{pathak_porosity_2025}.

\subsection{Ultrasonic inspection, mechanical and microstructural characterization}
\begin{figure}[htp!]
    \centering
    \includegraphics[width=\linewidth]{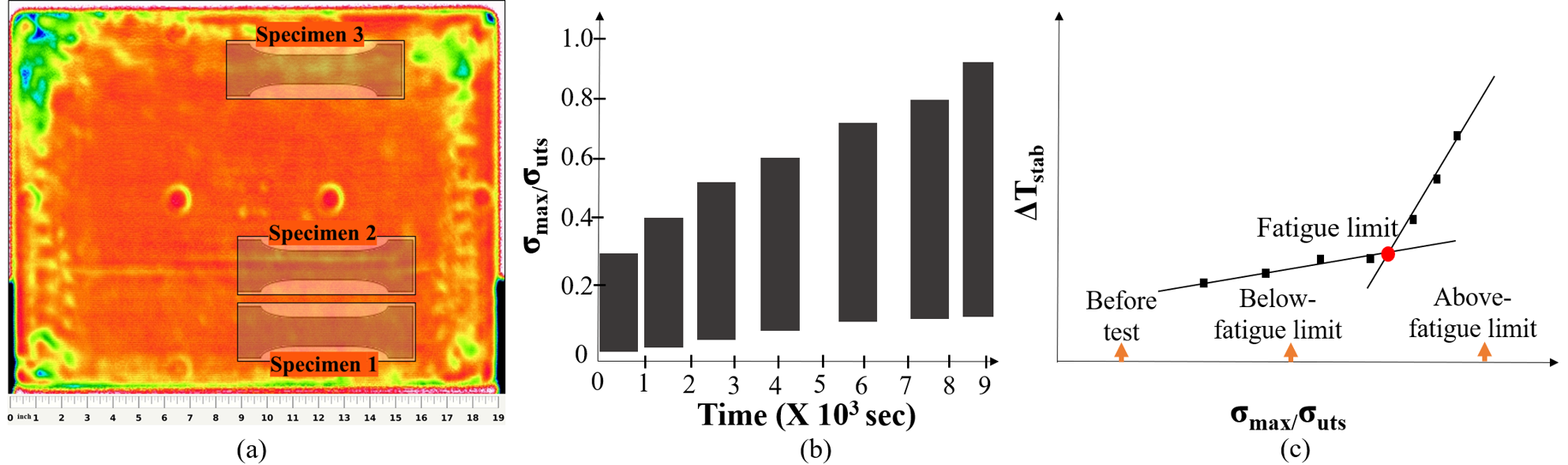}
    \caption{Specimen selection strategy and interrupted fatigue experimental protocol. (a) Ultrasonic C-scan attenuation map of the AM-CM panel used to select three specimens representing spatially distinct porosity levels; higher attenuation correlates with elevated local pore content. (b) Stepwise cyclic loading protocol showing progressive stress block increments periodically interrupted to allow thermal equilibration before subsequent loading. and c) Three $\mu$-CT imaging and quasi-static tensile testing stages: before fatigue testing (BT), below the fatigue limit (BF), and above the fatigue limit (AF), enabling direct correlation between evolving microstructure and degrading mechanical response within each individual specimen (adapted from \cite{pathak_porosity_2025}).}
    \label{fig:2}
\end{figure}

To investigate the influence of inherent microstructural heterogeneity on mechanical response, three representative specimens were selected from a single AM-CM panel following ultrasonic C-scan inspection (Figure~\ref{fig:2}a). The selection criterion was based on spatially varying ultrasonic attenuation signatures, which correlate with porosity content and local density variations. 

Each specimen was subjected to an interrupted fatigue protocol to capture microstructural evolution across progressively increasing damage states. Cyclic loading was applied in stepwise increments at a frequency of $15$ Hz and a stress ratio of $R = 0.1$, with each loading block comprising 7500 cycles at a prescribed stress amplitude. Starting from 30\% of ultimate tensile strength $(\sigma_{uts})$, the applied stress was increased incrementally in steps of 10\% $\sigma_{uts}$, up to specimen failure, as shown in Figure~\ref{fig:2}b. At each stress block, the specimen was cyclically loaded for the full 7500 cycles, after which it was interrupted and fully unloaded before either proceeding to the next increment or undergoing characterization. High-resolution $\mu$-CT scans were acquired at three key stages: the pristine state before any fatigue loading (BT), an intermediate stage below the fatigue limit at $50\%$ $\sigma_{uts}$ (BF), and a higher-load stage above the fatigue limit at $65\%$ $\sigma_{uts}$ (AF) shown in Figure~\ref{fig:2}c. At each interruption, the specimen was $\mu$-CT scanned and subsequently subjected to quasi-static tensile testing to quantify fatigue-induced stiffness degradation before reloading for the next increment.

$\mu$-CT imaging was performed using a Zeiss Xradia 620 Versa system at Auburn University. Specimens were mounted on a rotary stage perpendicular to the X-ray source, with a source-to-sample distance of 28 mm and a detector distance of 36 mm. Scans were conducted at 140 kV and 21 W using a 4$\times$ scintillator objective lens with 2$\times$2 binning and an air filter, yielding an isotropic voxel size of approximately 0.6~$\mu$m. Projection images were reconstructed into three-dimensional volumes using the system's built-in reconstruction software, producing approximately 1000 slices per specimen along the out-of-plane direction.

Quantitative microstructural characterization was performed by extracting geometric descriptors of fibers and pores from the reconstructed volumes. The fiber aspect ratio (AR) is defined as the ratio of fiber length to diameter ($l_f/d_f$). Fiber orientation is described by an azimuthal angle $\phi$ from the x-axis and a polar angle $\theta$ in the $y$--$z$ plane; the same convention is applied to pores. All descriptors were computed within 0.2~mm sub-volumes, a window size previously established to ensure convergence of constituent volume fractions and representative characterization of fiber and pore distributions~\cite{pathak_porosity_2025}. Descriptors were extracted using Dragonfly Pro software (version 2022.2), with further details provided in section \ref{sec:results}.

By repeatedly performing $\mu$-CT scanning and quasi-static testing on the same specimen at each stage, this approach establishes a direct correlation between evolving microstructural features and the corresponding degradation in mechanical properties, while eliminating specimen-to-specimen variability.

\subsection{Microstructure-resolved multiscale framework}
\begin{figure}[htb!]
    \centering
    \includegraphics[width=1\linewidth]{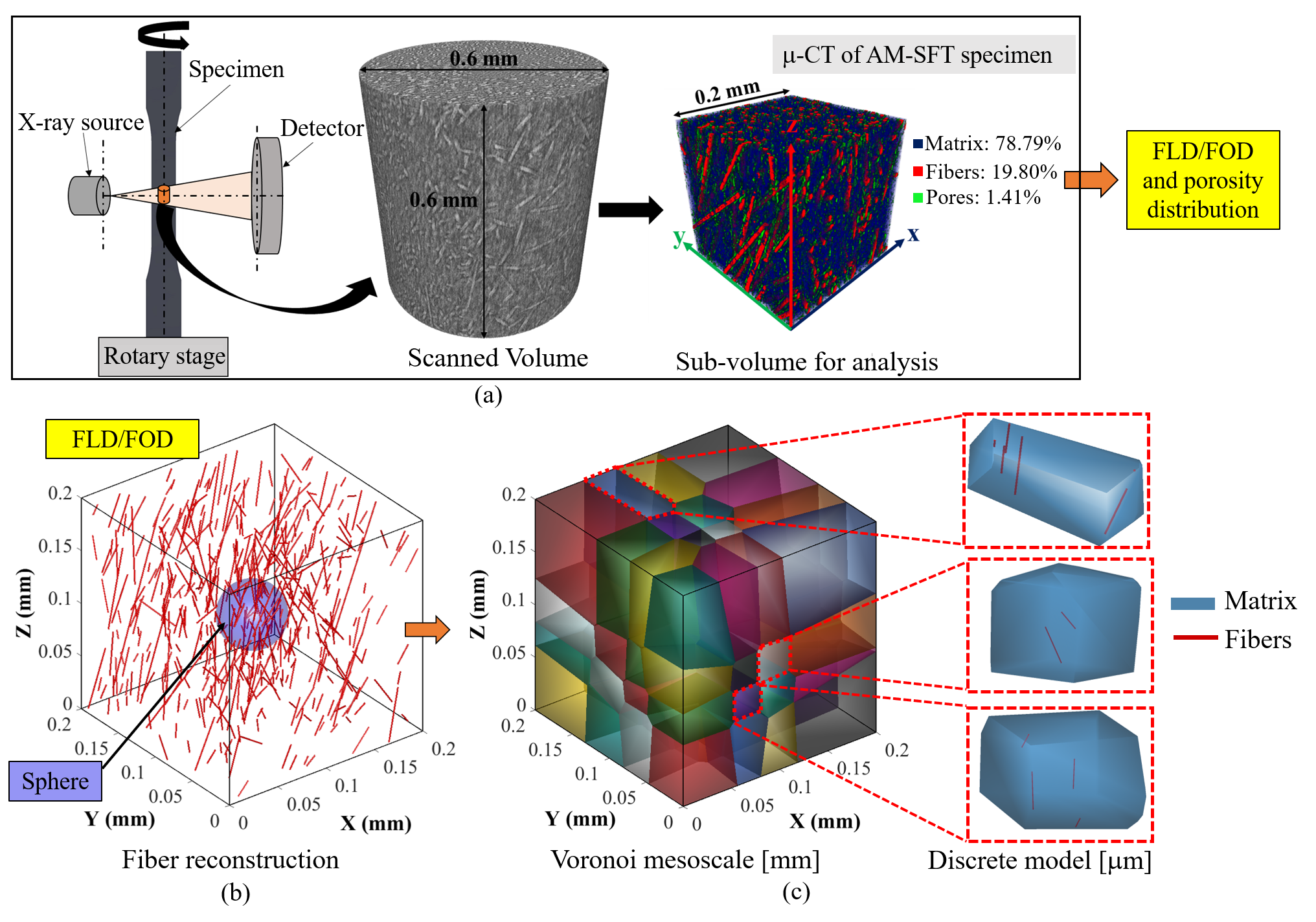}
    \caption{Microstructure reconstruction and mesoscale discretization workflow. (a) $\mu$-CT imaging of an AM-CM SFT specimen yields 3D-volumetric data segmented into carbon fiber, ABS matrix, and pore phases, from which fiber length, orientation, and clustering statistics are extracted within 0.2 mm sub-volumes. (b) A moving sphere traverses the reconstructed volume, grouping fibers whose centerlines intersect each sphere into localized interaction neighborhoods; representative points (centroids of fiber-sphere assemblies) serve as seeds for spatial partitioning. (c) Voronoi tessellation of the mesoscale cell produces a set of non-overlapping polyhedral cells, each representing a mechanically distinct fiber-interaction neighborhood for subsequent FE homogenization or surrogate evaluation (adapted from \cite{pathak_porosity_2025}).}
    \label{fig:3}
\end{figure}

Using microstructural geometric data obtained from $\mu$-CT, a microstructure-resolved multiscale framework has been developed to predict the quasi-static mechanical response of AM-CM SFTs. Figure~\ref{fig:3} presents the overall framework. To capture interaction physics while maintaining computational tractability at the coupon scale, discrete fiber-matrix nonlinear FE homogenization is replaced by a physics-informed surrogate model. This surrogate efficiently learns the mapping between fiber interaction topology and the strain-dependent stress response. Later sections describe each of these aspects in detail.

\subsubsection{Extracting fiber features from \texorpdfstring{$\mu$}{mu}-CT}
The volumetric data from high-resolution $\mu$-CT scans were segmented into three phases---carbon fibers, ABS matrix, and pores---as shown in Figure~\ref{fig:3}a. Fiber centerlines were then extracted from the segmented sub-volume (Figure~\ref{fig:3}b), preserving key geometric features governing mechanical response, including fiber lengths, orientation, clustering, and spatial distribution. In short-fiber thermoplastics, stress transfer is governed by shear-lag mechanisms, making the relative spatial arrangement of fibers critical. In particular, regions of high fiber density (clusters) and low fiber density (resin-rich zones) exhibit fundamentally different load transfer characteristics and damage evolution. Accurately preserving these spatial heterogeneities is therefore essential for any physically meaningful microstructural representation.

To capture these localized interactions, a moving-sphere sampling procedure is employed (Figure~\ref{fig:3}b). A sphere of radius $R_s$ traverses the reconstructed volume on a regular three-dimensional Cartesian grid with uniform spacing equal to $R_s$ in all three directions, and fibers whose centerlines intersect the sphere are grouped into local neighborhoods. The radius $R_s$ thus defines an interaction length scale: a smaller $R_s$ captures highly localized fiber interactions, while a larger $R_s$ yields coarser neighborhoods that average over multiple clusters and matrix-rich regions. Moreover, $R_s$ is selected as a global parameter based on the mean fiber length $l_{f,mean}$ obtained from the $\mu$-CT data. Specifically, $R_s \gtrsim l_{f,mean}$ ensures that individual fibers are sufficiently resolved within each sampled neighborhood and that shear-lag-based load transfer along the fiber length is not artificially truncated.

As the sphere traverses the microstructure, the chosen radius $R_s$ generates a set of overlapping neighborhoods that capture local fiber interactions. Due to this overlap, individual fibers may be included in multiple neighborhoods, ensuring continuity of interaction information across adjacent regions. Each neighborhood is associated with a representative point, defined as the centroid of the combined set comprising the moving sphere center and the centroids of all fibers contained within the sphere. These representative points provide a discrete sampling of the microstructure and form the basis for subsequent spatial partitioning. A visualization of the moving-sphere traversal across a representative microstructure is provided in Supplemental Material~\ref{sup:A}.

\subsubsection{Voronoi tessellation of mesoscale domain}
\label{sec:voronoi_tessellation}
\begin{figure}[htp!]
    \centering
    \includegraphics[width=1\linewidth]{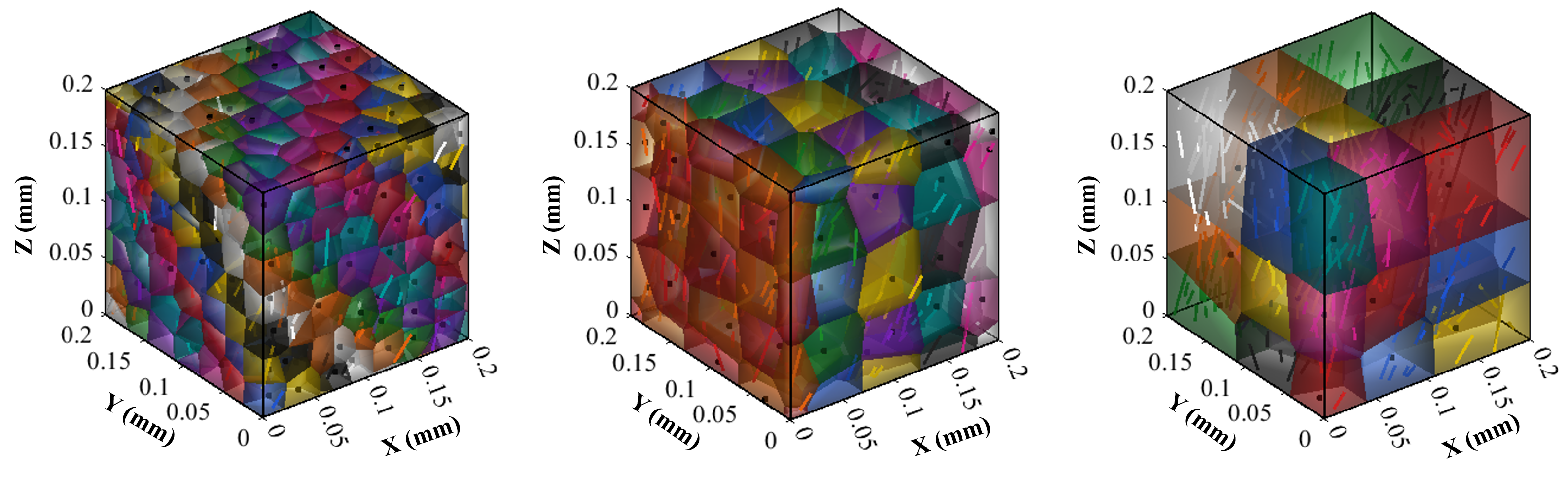}
    \caption{Dependence of Voronoi cell morphology on the moving-sphere sampling radius $R_s$, illustrated for three representative values: $R_s$ = 60 µm (left), $R_s$ = 120 µm (center), and $R_s$ = 180 µm (right). As $R_s$ increases, the number of seed points decreases, and individual cells grow larger.}
    \label{fig:4}
\end{figure}

The representative points obtained from the moving-sphere sampling procedure serve as seeds to construct a Voronoi tessellation of the mesoscale domain. This approach is preferred over alternative discretization methods, such as regular grids, $k$-means clustering, or Octree decomposition, because it preserves the intrinsic structural consistency of the underlying fiber distribution. First, because the seeds are derived directly from the fiber sampling process, the resulting cells are inherently topology-adaptive: fiber-dense regions naturally produce smaller cells, while matrix-rich regions yield larger cells, eliminating the need for user-defined mesh scales. Second, the tessellation provides a complete, non-overlapping partition of the domain, ensuring volumetric integrity during homogenization. Consequently, the mesoscale domain is decomposed into a set of non-overlapping polyhedral cells ($\Omega_c$) that collectively span the entire domain (Figure~\ref{fig:3}c). Third, the equidistant nature of Voronoi boundaries defines a geometrically consistent local neighborhood. This geometry is directly controlled by the sampling sphere radius ($R_s$), which dictates the minimum spacing between seeds and thereby influences the spatial distribution and local fiber volume fraction. For instance, a larger sampling radius ($R_s = 180~\mu\text{m}$) results in a coarser partition with fewer cells compared to a more refined sampling at $R_s = 60~\mu\text{m}$ (see Figure \ref{fig:4}). The Python implementation of the moving-sphere sampling and Voronoi tessellation described here is publicly available at \href{https://github.com/AMPL-Gururaja}{GitHub}~(\url{https://github.com/AMPL-Gururaja}). The repository includes scripts for sphere traversal, seed generation, Voronoi cell construction, STL export, and fiber volume fraction computation for each discrete cell.

\subsubsection{Discrete-cell Finite-Element Homogenization}
\label{Homogenization}
Each discrete Voronoi cell $\Omega_c$ is treated as a repeating unit cell (RUC) with periodic boundary conditions (PBCs) over which effective properties are evaluated using FE simulations. All FE simulations were performed using the commercially available software Abaqus 2022. Automated Python scripting was employed to generate equivalent cuboidal RUCs from the irregular Voronoi polyhedra through the following mapping procedure. 

First, the bounding box of each Voronoi cell is computed from its vertex coordinates, and an equivalent cuboid is constructed such that its volume matches the Voronoi cell volume. Second, the fiber centerline coordinates extracted from $\mu$-CT---specifically the start and end points of each fiber segment clipped to the Voronoi cell---are directly transferred into the cuboidal RUC coordinate system by applying a rigid translation that maps the Voronoi centroid to the cuboid centroid, preserving the absolute fiber orientations and inter-fiber spacing. This mapping ensures two key microstructural properties are preserved: (i) the fiber volume fraction within the Voronoi cell, and (ii) the fiber orientation distribution within the cell. 

The fibers and matrix were modeled using three-dimensional tetrahedral elements (C3D4), with PBCs applied on face, edge, and vertex pairs of the cuboidal RUC following the implementation detailed in our prior work \cite{Arabatti_2016, pathak_porosity_2025, LI_2004}. To extract the effective response of the RUC, six independent macroscopic strain cases (three normal and three shear) are applied sequentially in the form of far-field displacement fields. For each loading configuration, the stress and strain fields within the RUC are obtained from finite element simulations. The effective (homogenized) stress response is determined through volume averaging of the local fields at integration points using Equation \ref{eq:homogenization}:

\begin{equation}
\bar{\boldsymbol{\sigma}} = \frac{1}{V}\sum_{e=1}^{N_e}\sum_{g=1}^{N_{\mathrm{int}}} \boldsymbol{\sigma}^{g,e}\, V^{g,e}
\label{eq:homogenization}
\end{equation}

where $\bar{\boldsymbol{\sigma}}$ is the homogenized second-order stress tensor, $N_e$ is the total number of elements, $N_{\mathrm{int}}$ is the number of integration points per element (i.e.\ 1 in this case), $\boldsymbol{\sigma}^{g,e}$ is the stress tensor at the $g$-th integration point in the $e$-th element, $V^{g,e}$ is the volume (weight) of the $g$-th integration point in the $e$-th element, and $V$ is the total volume of the RUC.

The Voronoi cell-level effective response is represented by a strain-dependent constitutive operator (tangent stiffness evolution) $\mathbb{C}(\bar{\boldsymbol{\varepsilon}})$ and the corresponding stress--strain curve. Mesh convergence studies were conducted until the change in homogenized Young’s modulus between successive refinements fell below 2\%. A converged mesh element size of $0.5~\mu\text{m}$ was used in the remainder of the study. The effective stress–strain response obtained from FE homogenization of each discrete Voronoi cell is subsequently assembled across the full mesoscale Voronoi tessellation through volume-weighted averaging to predict the coupon-scale mechanical response, as described in detail in Section~\ref{assembly}.

\subsubsection{Constitutive Modeling of Constituents}
\label{sec:constitutive_constituents}
Carbon fibers are modeled as orthotropic linear-elastic solids using the properties listed in Table~\ref{Constituents properties}. Under the investigated quasi-static loading regime, post-mortem scanning electron microscopy (SEM) analysis reveals no evidence of fiber fracture \cite{pathak_porosity_2025}; stiffness degradation is therefore attributed primarily to matrix-driven degradation mechanisms, and fiber damage has not been accounted for in the current analysis. Additionally, the fiber-matrix interface is assumed to be perfectly bonded.

\begin{table}[h!]
\centering
\begin{threeparttable}
\renewcommand{\arraystretch}{1.2}
\caption{Elastic properties of the carbon fiber and ABS matrix constituents used in the 20 wt.\% CF-ABS discrete-cell models.}
\label{Constituents properties}
\begin{tabular}{|l|c|c|c|c|c|c|c|c|c|}
\hline
\textbf{Constituent} & \textbf{E$_{11}$} & \textbf{E$_{22}$} & 
\textbf{E$_{33}$} & \textbf{G$_{12}$} & \textbf{G$_{13}$} & 
\textbf{G$_{23}$} & \textbf{$\nu_{12}$} & \textbf{$\nu_{13}$} & 
\textbf{$\nu_{23}$} \\
& \textbf{(GPa)} & \textbf{(GPa)} & \textbf{(GPa)} & 
\textbf{(GPa)} & \textbf{(GPa)} & \textbf{(GPa)} & & & \\
\hline
Carbon fiber\textsuperscript{a} 
& 230 & 14 & 14 & 9 & 9 & 5 & 0.25 & 0.25 & 0.30 \\
\hline
ABS matrix\textsuperscript{b}   
& 2.31 & 2.31 & 2.31 & 0.86 & 0.86 & 0.86 & 0.35 & 0.35 & 0.35 \\
\hline
\end{tabular}
\begin{tablenotes}[flushleft]
    \footnotesize
    \item \textsuperscript{a} Carbon fiber properties adopted from literature \cite{WANG2013204}.
    \item \textsuperscript{b} ABS matrix elastic modulus and Poisson's ratio are determined experimentally.
\end{tablenotes}
\end{threeparttable}
\end{table}

The baseline elastic properties of the ABS matrix phase were assumed to be isotropic and were experimentally determined in this study by quasi-static uniaxial tensile testing in accordance with ASTM D638 Type I at 2 mm/min. Standalone neat ABS specimens were manufactured using the same processing conditions as the composite matrix phase. Testing was performed at room temperature under a displacement-controlled loading rate to isolate the baseline elastic modulus ($E = 2.31$~GPa) and Poisson's ratio ($\nu = 0.35$), which are assigned as the initial undamaged properties in Table~\ref{Constituents properties}.

\textbf{\textit{Matrix damage}}: The ABS matrix was modeled as an isotropic elastoplastic material with progressive damage to capture the experimentally observed nonlinearity, near-peak softening, and stiffness degradation. All damage model calibrations were performed using true stress--true strain data to ensure consistency under finite deformation. The initial elastic modulus $E$ was obtained via linear regression within the small-strain regime. The global yield stress $\sigma_Y$ was defined using the experimentally measured 0.2\% offset criterion, averaged across 3 specimens. 

\begin{figure}[htp!]
    \centering
\includegraphics[width=1\linewidth]{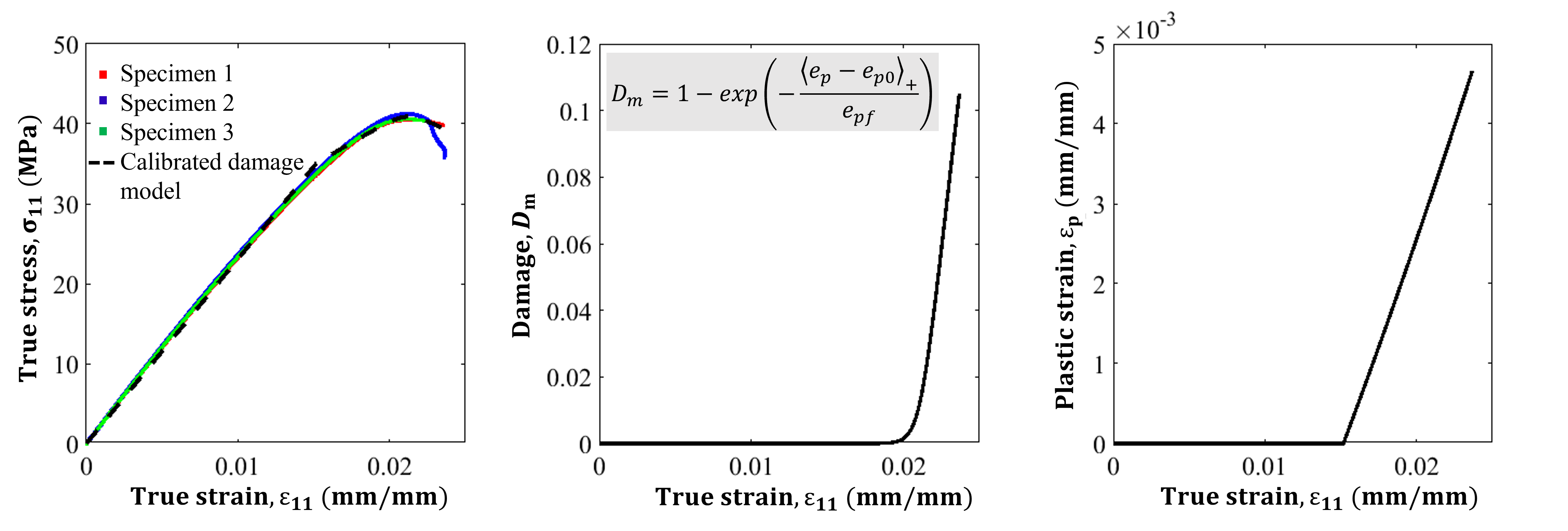}
    \caption{Calibration of the ABS matrix elastoplastic damage constitutive model. (a) Comparison of the calibrated two-term Voce plasticity plus exponential damage model against true stress $(\sigma_{11})$–true strain $(\varepsilon_{11})$ curves from three specimens, demonstrating excellent agreement through the elastic, nonlinear hardening, and post-peak softening regimes. (b) Evolution of the scalar damage variable $D_m$ with applied true strain $(\varepsilon_{11})$, showing damage initiation near peak stress followed by progressive softening. (c) Equivalent plastic strain $(\varepsilon_{p})$ at the onset of damage initiation (matrix failure index), illustrating the threshold beyond which crack-band-regularized softening is activated.}
    \label{fig:5}
\end{figure}

Plastic hardening was represented using a two-term Voce law \cite{polym15010234, Voce_1948}:

\begin{equation}
\sigma_y(e_p) = \sigma_Y+ R_1 \left(1 - e^{-b_1 e_p}\right)+ R_2 \left(1 - e^{-b_2 e_p}\right),
\label{eq:voce_law}
\end{equation}

where $e_p$ is the plastic strain and $R_1, b_1, R_2, b_2$ are global hardening parameters identified by fitting the true stress--strain response up to peak stress while enforcing consistency with the prescribed yield stress (see Figure~\ref{fig:5}a). The two-term form was adopted because ABS, as an amorphous thermoplastic, exhibits a characteristic saturating hardening response in which the hardening rate declines progressively with plastic strain \cite{Krairi2014}; a single exponential term is insufficient to simultaneously capture both the early nonlinear hardening and the near-peak plateau, whereas the two-term superposition provides the required flexibility without additional physical assumptions. 
Progressive stiffness degradation observed near and beyond peak stress was captured using a phenomenological scalar continuum damage mechanics (CDM) framework. Within this framework, the scalar damage variable $D_m \in [0,1)$ acts uniformly on the undamaged elastic stiffness tensor $\mathbb{C}_m$, so that all normal and shear stiffness components are degraded by the same factor $(1-D_m)$. The nominal Cauchy stress is therefore:
\begin{equation}
\boldsymbol{\sigma}_m = (1 - D_m)\,\mathbb{C}_m : \boldsymbol{\varepsilon}_m
\label{eq:effective_stress}
\end{equation}

This isotropic stiffness degradation representation is consistent with established CDM formulations for thermoplastic matrices \cite{Krairi2014}.

Damage initiates once $e_p \geq e_{p0}$ and evolves according to an 
exponential softening law (see Figure~\ref{fig:5}b and Figure~\ref{fig:5}c):
\begin{equation}
D_m = 1 - \exp\!\left(
    -\frac{\langle e_p - e_{p0}\rangle_{+}}{e_{pf}}
\right),
\label{eq:damage_law}
\end{equation}

where the Macaulay bracket $\langle\cdot\rangle_{+}$ ensures damage activates only when $e_p > e_{p0}$. The damage initiation strain $e_{p0}$ and the characteristic softening strain $e_{pf}$ were both treated as phenomenological parameters identified directly from the post-peak true stress--strain data of each specimen. The complete set of calibrated parameters is summarized in Table~\ref{tab:ABS_params}.

\begin{table}[htbp]
\centering
\caption{Calibrated ABS elastoplastic and damage parameters identified from true stress–strain data. Global plasticity parameters are common across all specimens; damage parameters are specimen-specific owing to differing porosity levels.}
\label{tab:ABS_params}
\begin{tabular}{lcc}
\hline
\textbf{Parameter} & \textbf{Symbol} & \textbf{Value} \\
\hline
Elastic modulus & $E$ & 2311.29 MPa \\
0.2\% offset yield stress & $\sigma_Y$ & 35.05 MPa \\
Voce parameter & $R_1$ & 13.37 MPa \\
Voce parameter & $b_1$ & 72.89 \\
Voce parameter & $R_2$ & 17.44 MPa \\
Voce parameter & $b_2$ & 72.73 \\
Damage initiation strain & $e_{p0}$ & 0.0032 \\
Characteristic softening strain & $e_{pf}$  & 0.013\\
\hline
\end{tabular}
\end{table}

\subsubsection{Effective Matrix Stiffness Incorporating Porosity}
\label{2-step homogenization}
\begin{figure}[htp!]
    \centering
    \includegraphics[width=1\linewidth]{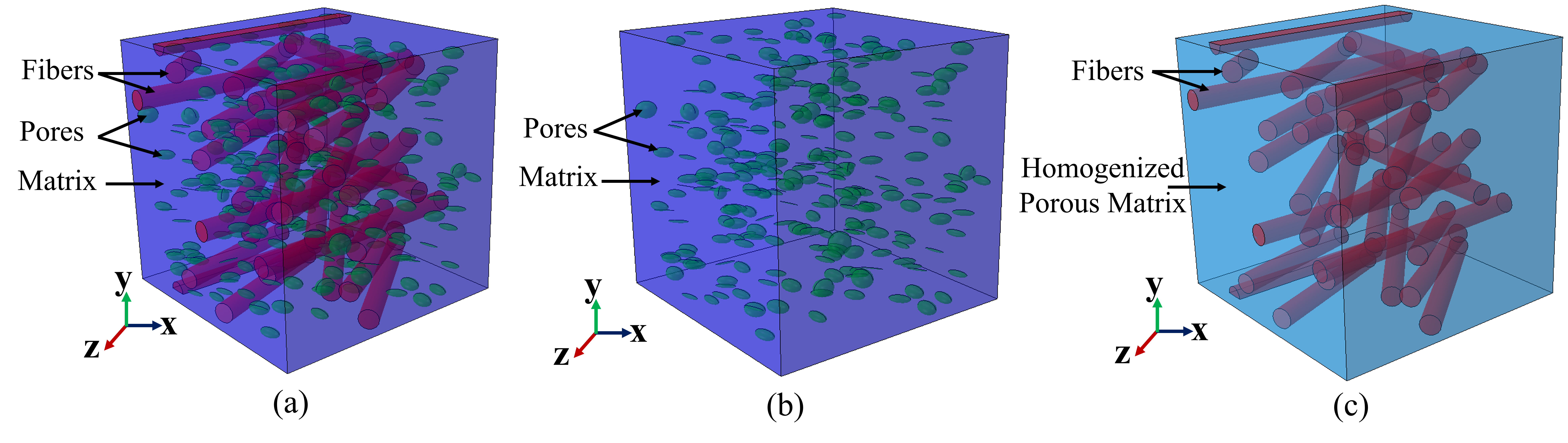}
    \caption{Figure 6:Two-step micromechanical homogenization procedure for incorporating matrix porosity effects. (a) Microstructure of SFTs consisting of carbon fibers and ABS matrix with microporosity, as reconstructed from $\mu$-CT imaging. (a) Step 1: a matrix-only RUC with $\mu$-CT-resolved pore morphology is homogenized under prescribed macroscopic strains to obtain the effective porous-matrix stiffness tensor  $C^{\text{eff}}_m$ ("Homogenized porous matrix"), (b) Step 2:  $C^{\text{eff}}_m$ is assigned to all fiber-matrix RUC used in the discrete-cell FE simulations ("Homogenized SFTs"), yielding fatigue-stage-resolved constitutive data that reflect measured porosity levels without explicitly meshing pores in every cell.}
    \label{fig:6}
\end{figure}

A two-step homogenization strategy was employed to incorporate the effect of porosity on the matrix's effective stiffness (Figure~\ref{fig:6}). First, a matrix-only RUC containing $\mu$-CT-resolved pores were numerically homogenized (similar to the procedure described in \ref{Homogenization}) to obtain an effective porous-matrix stiffness tensor $C^{\text{eff}}_m$.Second, this matrix stiffness tensor was assigned to all fiber-matrix RUC models. This procedure isolated pore-induced matrix degradation while allowing the fiber topology to vary spatially between cells \cite{Abhilash_2020}.

\subsection{Surrogate Modeling Framework}
\label{GNN_LSTM}
A surrogate constitutive framework was introduced to efficiently approximate the nonlinear, history-dependent homogenized response of each discrete Voronoi cell. For each Voronoi cell $\Omega_c$, the nonlinear finite-element homogenization defines a path-dependent constitutive operator in which the homogenized stress at pseudo-time $t$ depends on the entire applied macroscopic strain history up to $t$:
\begin{equation}
\bar{\boldsymbol{\sigma}}^{(\Omega_c)}(t) \;=\; \mathcal{H}_{\mathcal{T}_{\Omega_c}}\!\bigl(\{\bar{\boldsymbol{\varepsilon}}(s)\}_{s \le t}\bigr),
\label{eq:fe_operator}
\end{equation}

where $\{\bar{\boldsymbol{\varepsilon}}(s)\}_{s \le t}$ denotes the macroscopic strain history up to time $t$, and $\mathcal{H}_{\mathcal{T}_{\Omega_c}}$ is a constitutive operator parameterized by the local fiber-interaction topology $\mathcal{T}_{\Omega_c}$, comprising fiber geometry, spacing, clustering, and orientation, within $\Omega_c$. The explicit dependence on strain history reflects the path-dependent nature of the response, which arises from evolving internal variables (accumulated plastic strain $e_p$ and scalar damage $D_m$) in the elastoplastic-damage matrix described in Section~\ref{sec:constitutive_constituents}. The operator is therefore not reducible to a single-valued function of current strain, and any surrogate intended to replace it must inherit this history-dependent structure. As the number of discrete cells in a mesoscale assembly grew to the order of $10^{2}$--$10^{3}$, repeated evaluation of $\mathcal{H}_{\mathcal{T}_{\Omega_c}}$ through high-fidelity nonlinear FE simulations became computationally prohibitive. A surrogate operator $\hat{\mathcal{H}}_{\mathcal{T}_{\Omega_c}}$ is therefore constructed such that
\begin{equation}
\hat{\boldsymbol{\sigma}}^{(\Omega_c)}(t) \;=\; \hat{\mathcal{H}}_{\mathcal{T}_{\Omega_c}}\!\bigl(\{\bar{\boldsymbol{\varepsilon}}(s)\}_{s \le t}\bigr) \;\approx\; \mathcal{H}_{\mathcal{T}_{\Omega_c}}\!\bigl(\{\bar{\boldsymbol{\varepsilon}}(s)\}_{s \le t}\bigr).
\label{eq:surrogate_operator}
\end{equation}
The surrogate was trained on FE-generated stress--strain trajectories that satisfy equilibrium, the Hill--Mandel macrohomogeneity condition, and the two-term Voce isotropic-hardening elastoplastic-damage constitutive law calibrated for the ABS matrix phase (Section~\ref{sec:constitutive_constituents}). The construction of $\hat{\mathcal{H}}_{\mathcal{T}_{\Omega_c}}$ was decomposed into two complementary components that mirror the structure of classical history-dependent constitutive models: (i) a GNN that encodes the topology $\mathcal{T}_{\Omega_c}$ into a fixed-dimensional descriptor capturing the cell's stiffness potential and matrix-mediated load-transfer pathways, and (ii) a LSTM network that propagates the strain history through learned hidden- and cell-state vectors that serve as surrogate internal variables encoding accumulated plasticity and damage. The GNN therefore parametrizes the topology argument of $\hat{\mathcal{H}}$, while the LSTM parametrizes the history argument. These components are described in Sections~\ref{sec:gnn_encoder} and~\ref{sec:lstm_history}, respectively, and combined into the integrated architecture in Section~\ref{sec:integrated_arch}.

\subsubsection{Encoding Fiber Interactions via GNN}
\label{sec:gnn_encoder}
Each discrete Voronoi domain $\Omega_c$ was represented as an undirected weighted graph $\mathcal{G}_{\Omega_c} = (\mathcal{V}_{\Omega_c}, \mathcal{E}_{\Omega_c}, \mathcal{X}_{\Omega_c})$, where $\mathcal{V}_{\Omega_c}$ is the set of nodes representing individual fibers within the cell, $\mathcal{E}_{\Omega_c}$ is the set of weighted edges encoding shear-lag-mediated load-transfer pathways between fiber pairs, and $\mathcal{X}_{\Omega_c} \in \mathbb{R}^{N_f \times 8}$ is the node-feature matrix whose rows store per-fiber geometric attributes, with $N_f = |\mathcal{V}_{\Omega_c}|$ denoting the number of fibers in the cell. The microstructural input to the GNN for each fiber node $n$ consists of its three-dimensional start and end coordinates, $\mathbf{x}_n^{(s)}, \mathbf{x}_n^{(e)} \in \mathbb{R}^3$, extracted directly from $\mu$-CT, together with the cell-level fiber volume fraction $V_{f,\Omega_c}$ obtained from phase segmentation of the reconstructed volume (Figure~\ref{fig:3}). From these primitives, the fiber length $l_n = \| \mathbf{x}_n^{(e)} - \mathbf{x}_n^{(s)} \|$, orientation vector $\mathbf{t}_n = (\mathbf{x}_n^{(e)} - \mathbf{x}_n^{(s)})/l_n$, and per-fiber volume-fraction contribution $V_{f,n} = V_{f,\Omega_c} \, l_n / \sum_k l_k$ were computed for each Voronoi cell. To accelerate learning, an 8-dimensional feature vector is defined per fiber:

\begin{equation}
  \mathbf{x}_n =
  \left[
    l_n,\;
    V_{f,n},\;
    \mathbf{x}_n^{(s)},\;
    \mathbf{x}_n^{(e)}
  \right]
  \in \mathbb{R}^{8}
  \label{eq:node_features}
\end{equation}

\noindent

\begin{figure}[htp!]
    \centering
    \includegraphics[width=1\linewidth]{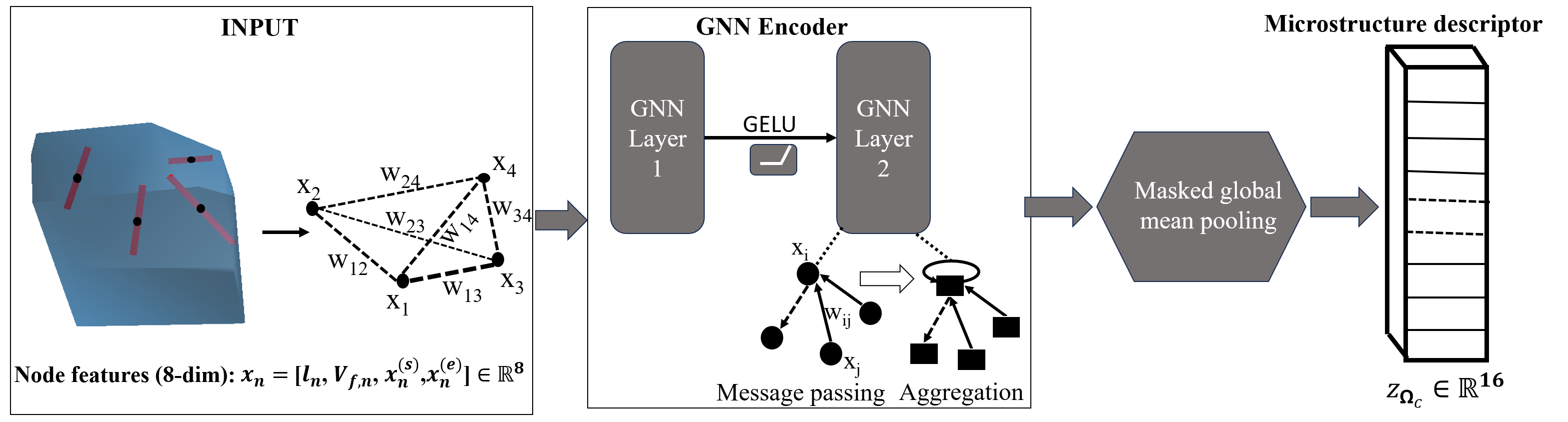}
    \caption{Graph construction and GNN encoding for a Voronoi cell. \textbf{Left}: Fibers are represented as graph nodes, with node features given by local volume fraction $V_{fiber,\Omega_c}$ and spatial endpoints $\mathbf{x}_n^{(s)}$,  $\mathbf{x}_n^{(e)}$. Fiber–fiber interactions were modeled as weighted edges $(w_{pq})$, computed from geometric proximity and directional alignment using a shear-lag-informed weight function (Equation \ref{edge_weight}). \textbf{Center}: The graph is processed through two stacked graph convolutional layers, where message passing and neighborhood aggregation update node embeddings to reflect local load redistribution. \textbf{Right}: Masked global mean pooling aggregates the final node embeddings into a compact 16-dimensional region-level microstructural descriptor $z_{\Omega_c} \in \mathbb{R}^{16}$, representing the combined effect of fiber geometry and interaction topology on the Voronoi cell's effective response.}
    \label{fig:7}
\end{figure}

Figure \ref{fig:7} depicts the graph representation for a discrete Voronoi cell. Here, each fiber node \textit{n} is represented using two descriptors: (i) the local fiber volume-fraction contribution $V_{f,n}$ and (ii) the spatial endpoint coordinates $\mathbf{x}_n^{(s)}$ and $\mathbf{x}_n^{(e)}$, which together encode the fiber length and orientation. The edges of $\mathcal{G}_{\Omega_c}$ are assigned weights $w_{pq}$ that quantify the load-transfer efficiency between fiber pairs (see Figure \ref{fig:7}).

\textbf{\textit{A note about edge weights}}: As previously mentioned, the edge weights capture the load-transfer efficiency between adjacent fiber pairs via matrix-mediated shear. In short-fiber composites, axial stress is transferred between adjacent fiber ends through interfacial shear stress transfer that occurs over a characteristic shear transfer length $L_c$. A neighboring fiber at end-to-end separation $d_{pq}$ therefore experiences a shear stress whose magnitude is proportional to $\exp(-d_{pq}/L_c)$: fiber pairs within $L_c$ interact strongly, while pairs separated by distances much greater than $L_c$ are effectively mechanically decoupled. $L_c$ was computed from Nairn's (1997) shear-lag formulation \cite{NAIRN199763, Lanids_1999, Cox_1952, Marshall_1985, Marshall_1987, Leon_2009}, which extended the classical Cox result to account for the geometric confinement of the matrix between neighboring fibers:

\begin{equation}
  L_c = r_f \sqrt{\frac{E^f_{11}\,\ln(R/r_f)}{2\pi\,G^m_{12}}}
  \label{eq:Lc}
\end{equation}

\noindent
where $r_f$ is the fiber radius, $E^f_{11}$ is the fiber axial Young's modulus (Table~1), $G^m_{12}$ is the matrix in-plane shear modulus (Table~1), and $R = r_f\sqrt{\pi/(2\sqrt{3}\,V_{f,\Omega_c})}$ is the mean matrix cylinder radius determined by local fiber packing geometry within the Voronoi cell.

The pairwise interaction weight between fibers $p$ and $q$ was defined as:

\begin{equation}
  w_{pq} =
  \exp\!\left(-\frac{d_{pq}}{L_c}\right)
  \!\left(0.90\,\lvert\mathbf{t}_p \cdot \mathbf{t}_q\rvert
         + 0.10\right)
  \label{edge_weight}
\end{equation}

\noindent
where $d_{pq}$ was the minimum segment-to-segment distance between fibers $p$ and $q$, and $\mathbf{t}_p$, $\mathbf{t}_q$ are their unit direction vectors. The exponential term encodes the attenuation of the shear-mediated stress field with fiber--fiber separation: as fiber ends move apart, the driving stress available to transfer load into a neighboring fiber diminishes exponentially over the length $L_c$.

The orientation-dependent edge weight is defined as $(0.90\lvert\mathbf{t}_p \cdot \mathbf{t}_q\rvert + 0.10)$, capturing both axial and residual interaction mechanisms. The dominant term (0.90) represents shear-lag-driven axial coupling, which scales with the cosine of the inter-fiber angle. It is strongest for parallel fibers, while highly misaligned fibers contribute minimally to axial stress redistribution. The constant term (0.10) accounts for a residual, orientation-independent interaction arising from transverse stress concentration due to fiber--matrix elastic stiffness mismatch, ensuring nonzero coupling even for orthogonal fibers. The $0.90/0.10$ split was determined from the normalized peak interfacial shear stress ratio obtained in the two-fiber interaction finite element simulations (Appendix~\ref{app:A1}). These simulations shown that the peak interfacial shear stress for orthogonal ($\phi = 90^\circ$) fibers dropped to approximately 10\% (ranging between 8--22\% depending on the inter-fiber gap distance) of the fully aligned ($\phi = 0^\circ$) baseline configuration. This baseline residual ratio served as the basis for using 0.10, while the remaining normalized load-bearing capacity (0.90) was allocated to the orientation-dependent cosine term.

The edge weight function operates strictly at the pairwise level. Higher-order multi-fiber interaction effects---such as load redistribution within fiber clusters, and cooperative load sharing---are not explicitly encoded in the edge weights but emerge through neighborhood aggregation in the GNN message-passing operation (Equation~\eqref{eq:message_passing}). The graph is undirected $w_{pq} = w_{qp}$ and complete: every pair of fibers within a Voronoi cell is connected. A lower bound of $10^{-6}$ was imposed on edge weights to maintain numerical stability of the degree-normalized aggregation in Equation~\eqref{eq:message_passing}, without altering the graph topology. The parametric finite-element study (Appendix~\ref{app:A1}) further confirmed that both the cosine-dependent axial coupling and the orientation-independent residual are physically operative for fiber geometries and spacings consistent with $\mu$-CT observations.

\textbf{\textit{Graph convolutional network (GCN) layer formulation}}: Node features were first projected from the 8-dimensional input space to a 64-dimensional embedding space via a linear layer with GELU activation. Two stacked graph convolutional layers then updated node embeddings through degree-normalized weighted aggregation with residual connections:

\begin{equation}
  \mathbf{h}_n^{(l+1)} =\mathbf{h}_n^{(l)} +  \mathrm{Dropout}\!\left(\mathrm{LN}\!\left(\mathrm{GELU}\!\left(\sum_{m \in \mathcal{N}(n)}\frac{w_{nm}}{\sqrt{\hat{d}_n\,\hat{d}_m}}\,\mathbf{W}^{(l)}\,\mathbf{h}_m^{(l)}\right)\right)\right)
  \label{eq:message_passing}
\end{equation}

\noindent
where $\mathbf{h}_m^{(l)}$ is the embedding of neighboring fiber node $m$ at layer $l$, $w_{nm}$ is the edge weight from Equation~\ref{edge_weight}, $\hat{d}_n = 1 + \sum_j w_{nj}$ is the weighted node degree with self-loop, $\mathbf{W}^{(l)}$ is the trainable weight matrix at layer $l$, LN denotes layer normalization, and the residual connection $\mathbf{h}_n^{(l)}$ stabilized gradient flow across layers. The degree normalization factor $1/\sqrt{\hat{d}_n\hat{d}_m}$ prevents embedding magnitudes from scaling with local fiber connectivity, ensuring that fibers in dense clusters were not artificially amplified relative to isolated fibers. From a mechanical standpoint, this message-passing operation approximated multi-fiber load redistribution within the fiber network: edge weights encoded pairwise shear-mediated stress-transfer efficiency, neighborhood aggregation represented multi-fiber load sharing among all neighboring fibers, and stacked layers iteratively propagated this redistribution across the interaction network---analogous to FE solver iterations.

After two message-passing layers, masked global mean pooling aggregated active node embeddings into a 64-dimensional graph-level representation, which was projected by a linear layer to a compact 16-dimensional microstructural descriptor:

\begin{equation}
  z_{\Omega_c}
  =
  \mathbf{W}_{\mathrm{pool}}\!
  \left(
    \frac{1}{N_f}
    \sum_{n=1}^{N_f} \mathbf{h}_n^{(2)}
  \right)
  \in \mathbb{R}^{16}
  \label{eq:pooling}
\end{equation}

\noindent
This embedding served as a compact, fixed-length representation of the cell's effective stiffness potential and interaction topology,
capturing the combined influence of fiber geometry, spatial distribution, volume fraction, and interaction network structure on the cell's effective mechanical response. Because $z_{\Omega_c}$ was derived exclusively from quantities computable from $\mu$-CT fiber coordinates and $V_{f,\Omega_c}$, the GNN required no microstructural inputs beyond standard reconstruction outputs.

\subsubsection{Modeling History-Dependent Nonlinear Evolution via LSTM}
\label{sec:lstm_history}
The homogenized stress response of each cell was a history-dependent trajectory that reflected progressive accumulation of matrix plastic
strain $e_p$ and scalar damage $D_m$ under monotonic loading. Because these internal variables evolved with loading history, the
stress at any increment depended not only on the current strain but on the entire prior deformation path. A feedforward network that treats each strain increment independently could not enforce this temporal consistency. A LSTM network was therefore employed to process the loading history as a sequence, maintaining hidden state vectors $\mathbf{h}_t \in \mathbb{R}^{128}$ and cell state vectors $\mathbf{c}_t \in \mathbb{R}^{128}$ as surrogate internal state variables implicitly encoding accumulated plasticity and damage without explicitly tracking them.

\textit{\textbf{LSTM input features}}: The LSTM input at each strain increment $t$ was a 21-dimensional feature vector concatenated from two categories of inputs:

\begin{equation}
  \mathbf{q}_t =
  \underbrace{
    \left[
      \frac{\varepsilon_t}{\varepsilon_{\mathrm{pk}}},\;
      \left(\frac{\varepsilon_t}{\varepsilon_{\mathrm{pk}}}\right)^{\!2},\;
      \left(\frac{\varepsilon_t}{\varepsilon_{\mathrm{pk}}}\right)^{\!3},\;
      \ln\!\left(1+\frac{\varepsilon_t}{\varepsilon_{\mathrm{pk}}}\right),\;
      V_{f,\Omega_c}
    \right]
  }_{\text{5 loading-side features}}
  \;\Big\|\;
  \underbrace{z_{\Omega_c}}_{\text{16 D topology}}
  \label{eq:lstm_input}
\end{equation}
\noindent
These input vector maintains a fixed size of 21 dimensions across all strain increments rather than growing with the sequence length. The history of prior deformation is captured entirely through the step-by-step updates of the fixed-dimensional internal state.
The first category comprised five loading-side features derived analytically from the applied scalar strain $\varepsilon_t$ and the
fiber coordinate inputs used by the GNN. The polynomial and logarithmic strain features, motivated by the hyperelastic model, enriched the nonlinear strain representation, enabling the LSTM to distinguish between the linear elastic, nonlinear hardening, and near-peak softening regimes without requiring an explicit constitutive form. Normalizing strain by $\varepsilon_{\mathrm{pk}}$ mapped the stress peak of every cell to a universal reference strain of 1.0, enabling the LSTM to learn a shape that was subsequently amplitude-rescaled. The cell-level volume fraction $V_{f,\Omega_c}$ directly conveyed the available reinforcement content.

The LSTM hidden and cell state vectors evolved as:
\begin{equation}
  \left(\mathbf{h}_t,\,\mathbf{c}_t\right)
  =
  \mathrm{LSTM}\!\left(\mathbf{q}_t,\;
                       \mathbf{h}_{t-1},\;
                       \mathbf{c}_{t-1}\right)
  \label{eq:lstm}
\end{equation}

\noindent
with two stacked layers (hidden dimension 128, inter-layer dropout 0.25). The hidden and cell states serve as surrogate internal variables implicitly encoding accumulated plastic strain and damage history at each loading increment. The GNN component performed geometric and topological encoding of the fiber interaction network, while the LSTM modeled the path-dependent evolution of the constitutive response --- a separation that mirrors classical constitutive modeling, in which microstructural descriptors govern stiffness capacity while evolution laws govern nonlinear progression.

\subsubsection{Integrated GNN--LSTM Architecture}
\label{sec:integrated_arch}
\begin{figure}[htp!]
    \centering
    \includegraphics[width=1\linewidth]{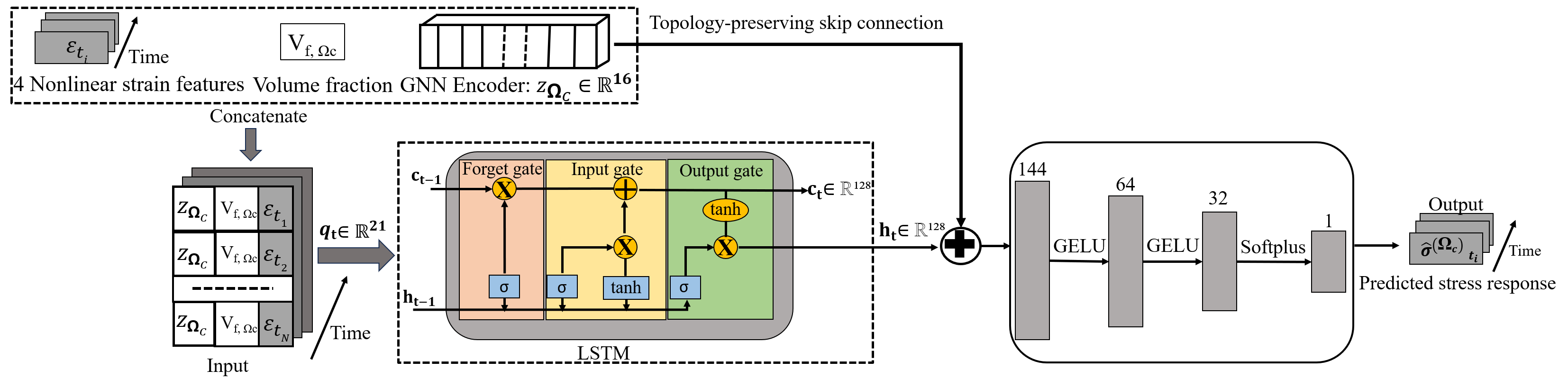}
    \caption{Integrated GNN--LSTM architecture for physics-informed, history-dependent stress--strain prediction. The GNN microstructure encoder (left) processes the fiber interaction graph $\mathcal{G}_{\Omega_c}$ to produce a fixed 16-dimensional cell embedding $z_{\Omega_c} \in \mathbb{R}^{16}$. At each strain increment $t$, this embedding is concatenated with four nonlinear strain features and the cell-level volume fraction $V_{f,\Omega_c}$ to form the 21-dimensional input vector $\mathbf{q}_t \in \mathbb{R}^{21}$, which is fed sequentially into a two-layer LSTM (hidden dimension 128). The LSTM hidden state $\mathbf{h}_t \in \mathbb{R}^{128}$ evolves as a surrogate set of internal state variables encoding accumulated plasticity and damage history. A skip connection concatenates $z_{\Omega_c}$ directly with $\mathbf{h}_t$, producing a 144-dimensional input to a three-layer output MLP ($144 \rightarrow 64 \rightarrow 32 \rightarrow 1$, Softplus), which yields the normalized predicted stress $\hat{\sigma}_t^{(\Omega_c)}$ at each increment.}
    \label{fig:8}
\end{figure}

The final surrogate mapping from fiber interaction topology to normalized stress trajectory was:
\begin{equation}
  \hat{\sigma}_t^{(\Omega_c)}
  =
  \mathrm{Softplus}\!\left(
    \mathcal{F}_{\mathrm{out}}\!\left(
      \left[\mathbf{h}_t\;;\;z_{\Omega_c}\right]
    \right)
  \right),
  \quad
  \mathbf{h}_t
  =
  \mathcal{F}_{\mathrm{LSTM}}\!\left(
    \mathbf{q}_t,\,\mathbf{h}_{t-1},\,\mathbf{c}_{t-1}
  \right),
  \quad
  z_{\Omega_c}
  =\mathcal{F}_{\mathrm{GNN}}\!\left(\mathcal{G}_{\Omega_c}\right)
  \label{eq:surrogate_mapping}
\end{equation}

\noindent
where $\mathcal{F}_{\mathrm{GNN}}(\mathcal{G}_{\Omega_c})$ denoted the GNN encoder operating on the fiber interaction graph
$\mathcal{G}_{\Omega_c}$ of cell $\Omega_c$, and $\mathcal{F}_{\mathrm{out}}$ was a three-layer residual output network.
Figure~\ref{fig:8} illustrates the complete architecture.

The output network received a skip-connected input formed by concatenating the LSTM hidden state
$\mathbf{h}_t \in \mathbb{R}^{128}$ with the GNN embedding $z_{\Omega_c} \in \mathbb{R}^{16}$ at every strain increment,
producing a 144-dimensional input to a three-layer MLP ($144 \to 64 \to 32 \to 1$, GELU activations).
The skip connection ensured that the GNN-encoded topology signal---which carried direct information about the cell's peak stress potential through its fiber arrangement---remained accessible at every prediction step independently of LSTM hidden state depth.
Without this connection, the GNN amplitude signal would need to propagate through all LSTM gates across potentially 100 or more time steps, risking attenuation through gate saturation. A Softplus nonlinearity, defined as $f(x) = \ln(1 + e^{x})$, at the network output enforced the physical constraint that predicted stress remained strictly non-negative throughout the trajectory.

To recover stress in physical units, the surrogate output was trained against FE-computed stress trajectories normalized by the cell-level peak stress, with the peak stress predicted independently from the GNN embedding via a post-hoc regression model trained on the same FE dataset. This architecture enforced a physically meaningful factorization: the GNN captured topology-dependent stiffness potential from the fiber interaction structure; the LSTM captured the path-dependent evolution of the constitutive response.

The peak stress amplitude was predicted using a Random Forest regressor (300 trees, MultiOutputRegressor) trained post-hoc on the frozen GNN embedding $z_{\Omega_c}$ concatenated with the cell-level volume fraction $V_{f,\Omega_c}$, using the same 500-cell dataset and 70--15--15 train--validation--test split. Although the GNN auxiliary head also predicted peak stress during joint training, it exhibited generalization degradation on unseen cells due to its coupling with the evolving LSTM gradients during end-to-end optimization. The post-hoc Random Forest, applied to frozen embeddings after GNN--LSTM convergence, provided more robust amplitude generalization without requiring additional finite-element simulations, achieving a coefficient of determination $R^2 = 0.98$ on the held-out test set for peak stress prediction compared with the discrete FE model.

\subsubsection{Physics-Informed Regularization}
\textit{\textbf{Composite loss function}}: To ensure physically plausible surrogate predictions, a composite loss function was used alongside additional structural training constraints. Model parameters were optimized by minimizing:
\begin{equation}
  \mathcal{L}
  =
  \mathcal{L}_{\mathrm{Huber}}
  + 0.05\,\mathcal{L}_{\mathrm{mono}}
  + 0.001\,\mathcal{L}_{\mathrm{smooth}}
  + 0.02\,\mathcal{L}_{\mathrm{zero}}
  + 0.3\,\mathcal{L}_{\mathrm{aux}}
  + 0.5\,\mathcal{L}_{\mathrm{peak}}
  \label{eq:composite_loss}
\end{equation}
\noindent
The scalar scaling coefficients were assigned to each loss component (0.05, 0.001, 0.02, 0.3, and 0.5) and function as regularization hyperparameters. These constants were determined empirically via a grid search on the validation subset to ensure stable convergence during joint training, balancing trajectory-tracking accuracy with physical admissibility constraints (e.g., preventing non-physical oscillations without over-penalizing post-yield softening).The primary term $\mathcal{L}_{\mathrm{Huber}}$ was a strain-weighted Huber loss ($\delta = 0.1$) computed exclusively over real,non-padded data points:

\begin{equation}
  \mathcal{L}_{\mathrm{Huber}}
  =
  \frac{
    \displaystyle\sum_t
      w_t\,\mathcal{H}_{\delta}\!\left(
        \hat{\sigma}_t - \sigma_t^{\mathrm{FE}}
      \right)
  }{
    \displaystyle\sum_t w_t
  }
  \label{eq:huber_loss}
\end{equation}

\noindent
where $\hat{\sigma}_t$ denotes the surrogate-predicted normalized stress at strain increment $t$, $\sigma_t^{\mathrm{FE}}$ represents the corresponding high-fidelity FE homogenization target stress, and $w_t$ is the strain-dependent tracking weight matrix. The tracking weights are defined explicitly as:

\begin{equation}
  w_t
  =
  1
  +
  3\!\left[
    \mathrm{clip}\!\left(
      \frac{\varepsilon_t/\varepsilon_{\mathrm{pk}} - 0.8}{0.8},\;
      0,\;1
    \right)
  \right]^{\!2}
  \label{eq:strain_weights}
\end{equation}
\noindent
where $\varepsilon_t$ is the applied macroscopic strain at step $t$, and $\varepsilon_{\mathrm{pk}}$ is the localized cell-level strain corresponding to the peak stress point. The term $\mathrm{clip}(x, a, b)$ represents a standard mathematical bounding operation defined to constrain the scaling window strictly between $a$ and $b$.

This custom weighting scheme is engineered to prioritize tracking accuracy along highly nonlinear paths. It assigns a tracking penalty up to $4\times$ higher to data points approaching and exceeding the stress peak ($\varepsilon_t/\varepsilon_{\mathrm{pk}} \geq 0.8$), where accurate prediction of the onset and progression of nonlinear matrix softening is most mechanically significant. The Huber loss component $\mathcal{H}_{\delta}$ is intentionally employed in preference to a standard mean-squared error framework to reduce gradient sensitivity to extreme outlier stress fluctuations arising from highly localized damage events in individual cells, while retaining stable quadratic penalization for minor prediction errors.

\textit{\textbf{Pre-peak monotonicity}}: Stress increments prior to the peak were regularized to prevent non-physical oscillations:

\begin{equation}
  \mathcal{L}_{\mathrm{mono}}
  =
  \frac{1}{t_{\mathrm{pk}}-1}
  \sum_{t=1}^{t_{\mathrm{pk}}-1}
    \mathrm{GELU}\!\left(
      \hat{\sigma}_{t-1} - \hat{\sigma}_t
    \right)
  \label{eq:mono_loss}
\end{equation}

\noindent
where $t_{\mathrm{pk}}$ was the index of the true stress peak identified from the FE data for each cell. The post-peak regime was explicitly excluded from this constraint, consistent with the experimentally observed and FE-computed post-peak softening behavior of the ABS matrix following peak stress.

\textbf{\textit{Smoothness:}} Second-order differences of the predicted stress trajectory were penalized to suppress artificial stiffness fluctuations:

\begin{equation}
  \mathcal{L}_{\mathrm{smooth}}
  =
  \frac{1}{T-2}
  \sum_{t=1}^{T-2}
    \left(
      \hat{\sigma}_{t+1}
      - 2\hat{\sigma}_t
      + \hat{\sigma}_{t-1}
    \right)^{\!2}
  \label{eq:smooth_loss}
\end{equation}
\noindent
where $T$ represents the total sequence length (total number of strain increments) in the complete loading history..

\textbf{\textit{Elastic consistency:}} The predicted normalized stress at the first strain increment was penalized to enforce physically correct onset from zero stress:
\begin{equation}
  \mathcal{L}_{\mathrm{zero}}
  =
  \hat{\sigma}_0^{\,2}
  \label{eq:zero_loss}
\end{equation}
\noindent
where $\hat{\sigma}_0$ is the surrogate-predicted normalized stress at the initial strain increment ($t=0$), ensuring that the predicted stress trajectory originates strictly from a state of zero stress.

\textbf{\textit{Auxiliary physics supervision:}} The auxiliary loss $\mathcal{L}_{\mathrm{aux}}$ penalized deviation of the GNN auxiliary head predictions $[\tilde{E},\tilde{\sigma}_{\mathrm{pk}}, \tilde{\varepsilon}_{\mathrm{pk}},\tilde{\delta}]$ from their normalized FE-computed targets, ensuring that the 16-dimensional cell embedding encoded mechanically interpretable information---initial modulus, peak stress, peak strain, and post-peak softening rate---throughout training rather than converging to a representation that satisfied the trajectory loss without physical meaning.

\textbf{\textit{Peak stress penalty}}: The term:

\begin{equation}
\mathcal{L}_{\text{peak}} = \left\| \hat{\mathbf{\sigma}}_{t_{\mathrm{pk}}} - \mathbf{\sigma}^{\text{FE}}_{t_{\mathrm{pk}}} \right\|^2
  \label{eq:peak_loss}
\end{equation}
\noindent
applied concentrated penalization at the exact peak stress point, improving prediction accuracy at the mechanically critical transition
between hardening and softening regimes independently of the weighted Huber loss.

\textit{\textbf{Optimizer and training schedule}}: The GNN and LSTM were trained jointly end-to-end using the AdamW optimizer (initial learning rate $5 \times 10^{-4}$, weight decay $2 \times 10^{-4}$) \cite{AdamW} with gradient norm clipping to 1.0 to prevent unrealistically high gradients during early training epochs. A ReduceLROnPlateau learning rate scheduler \cite{Bengio2012} halved the learning rate when the validation loss plateaued for 15~consecutive epochs, with a minimum learning rate of $10^{-6}$. Training ran for up to 800 epochs; early stopping was applied based on validation loss, with a patience of 57 epochs, to prevent overfitting. The complete set of training hyperparameters is summarized in Appendix~\ref{app:A2}.

\subsubsection{Error Propagation and Stability}
Small-cell-level prediction errors with the GNN-LSTM surrogate did not lead to any observed instabilities in the coupon-scale simulations. Because the coupon-level response was obtained by volume averaging over many discrete cells, surrogate approximation errors were statistically mitigated through the averaging operation itself. Robust macroscopic predictions were therefore achievable even in the presence of small cell-level surrogate approximation errors, with the principal sources of coupon-scale discrepancy attributable to structural idealizations in the Voronoi representation rather than to surrogate approximation errors.

\subsection{Mesoscale Assembly and Coupon-Scale Prediction}
\label{assembly}
Having obtained the effective stress--strain response of each discrete Voronoi cell---either through direct FE homogenization (Section~\ref{Homogenization}) or through the GNN-LSTM surrogate (Section~\ref{GNN_LSTM})---the collection of cell responses defines a heterogeneous mesoscale constitutive field over the reconstructed volume. The coupon-scale response was obtained by assembling the Voronoi cell responses through volume-weighted averaging, consistent with the Hill--Mandel homogenization principle:
\begin{equation}
\mathbf{\Sigma}(\mathbf{\epsilon}) = \frac{1}{V_{\text{coupon}}} \sum_{\Psi_{c}} V_{\Psi_{c}} \bar{\mathbf{\sigma}}^{(\Psi_{c})}(\mathbf{\epsilon})
\end{equation}
where $\mathbf{\Sigma}$ is the macroscopic second-order stress tensor of the coupon, $V_{\text{coupon}}$ is the total volume of the coupon, $\Psi_c$ denotes the $c$-th mesoscale Voronoi cell within the reconstructed assembly, $V_{\Psi_c}$ is the volume of that corresponding cell, and $\bar{\mathbf{\sigma}}^{(\Psi_{c})}$ is the homogenized second-order stress tensor predicted by the trained GNN-LSTM surrogate predictions under the applied macroscopic strain tensor $\epsilon$. 

Because the macroscopic response is obtained by aggregation over many cells, moderate cell-level surrogate errors are naturally stabilized through averaging. This enables robust coupon-scale prediction while preserving sensitivity to the spatial distribution of fiber interaction topology.

\section{Results and Discussion}
\label{sec:results}
\subsection{Microstructural Evolution under Cyclic Loading}
\begin{figure}[htb!]
    \centering
    \includegraphics[width=1\linewidth]{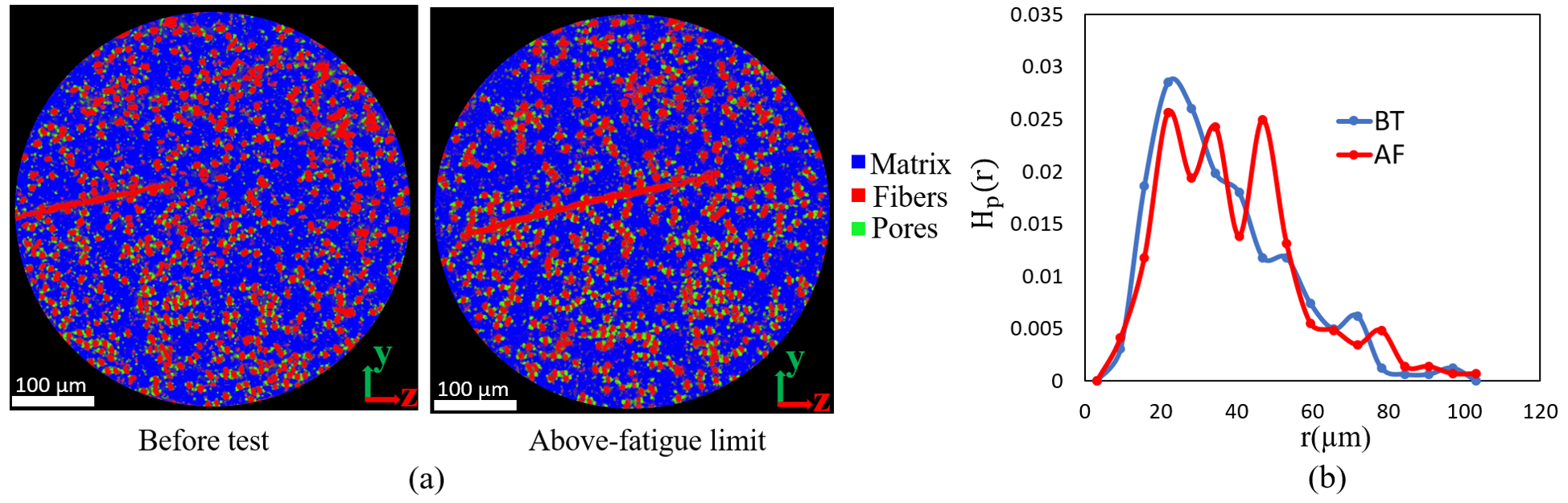}
    \caption{Fatigue-driven microstructural damage in 20 wt.\% CF-ABS, fabricated via AM-CM at ORNL, was characterized using $\mu$-CT. (a) Representative cross-sectional images at two damage states are shown: prior to fatigue loading (Before test) and after exceeding the fatigue limit (Above-fatigue limit). Post-fatigue scans reveal pronounced localized pore nucleation and growth, predominantly near fiber termini and within fiber-clustering regions, indicating stress-concentration-mediated damage initiation and propagation. (b) Nearest-neighbor probability density $H_p(r)$ tracked longitudinally for a representative sample (Specimen 2) for the Before test (BT) and Above fatigue limit (AF) stages, indicating progressive cluster dilation driven by pore growth within the inter-fiber zones, resulting in a dispersed sub-population of fibers with substantially increased nearest-neighbor distances.}
    \label{fig:9}
\end{figure}

$\mu$-CT analysis revealed that pore nucleation and growth were highly localized. Cyclic loading induced measurable shifts in ensemble-level microstructural descriptors, with pores forming preferentially in (i) fiber-end neighborhoods and (ii) clustered fiber regions where local matrix ligaments are thin (Figure~\ref{fig:9}a). These locations coincide with regions of elevated shear and normal stress concentration as predicted by shear-lag theory and inter-fiber load-transfer mechanics. The preferential localization of porosity near fiber termini provided direct experimental evidence that stress-transfer discontinuities at fiber ends serve as primary damage-initiation sites.

The nearest-neighbor probability density $H_p(r)$ provides complementary quantitative evidence of progressive fiber-cluster dispersion accompanying pore growth (Figure~\ref{fig:9}b). Before testing, $H_p(r)$ exhibited a single dominant peak near $r \approx 20$--$22~\mu\text{m}$, consistent with tight fiber clustering inherited from the manufacturing process. Above the fatigue limit, however, a secondary peak emerged near $r \approx 45$--$47~\mu\text{m}$, producing a distinct bimodal distribution that signifies a clear signature of cluster dilation. 

While fiber rotation is geometrically constrained within a pristine, continuous polymer matrix, this observable shift in the FOD is a direct structural consequence of severe, localized matrix failure that accumulates during cyclic loading. As fatigue cycles progress, a massive threefold increase in localized matrix porosity occurred preferentially within the thin matrix ligaments separating dense fiber clusters and around stress-concentrating fiber ends (see Table~\ref{tab:porosity_volume}). This progressive micro-voiding, debonding, and matrix degradation effectively dismantled the local geometric constraints holding the fibers rigid. Under macroscopic tensile strain, the fibers adjacent to these expanding void networks are physically tilted and reallocated into the newly vacated micro-volumes. This localized rearrangement and shifting of the fiber network under cyclic loading directly aligns with experimental observations of damage-induced microstructural evolution reported in the literature ~\cite{arif_multiscale_2014, Almeida}.

\begin{table}[htb!]
\centering
\caption{Porosity volume fraction (\%) evolution under cyclic loading. Values represent the mean ± standard deviation (adopted from \cite{pathak_porosity_2025}).}
\label{tab:porosity_volume}
\begin{tabular}{c c c c}
\hline
\textbf{Specimen} & \textbf{Before test} & \textbf{Below fatigue limit} & \textbf{Above fatigue limit} \\
\hline
1 & $1.41 \pm 0.17$ & $2.33 \pm 0.21$ & $6.63 \pm 0.70$ \\
2 & $1.90 \pm 0.15$ & $2.56 \pm 0.18$ & $8.64 \pm 0.60$ \\
3 & $2.86 \pm 0.11$ & $3.57 \pm 0.15$ & $11.86 \pm 0.56$ \\
\hline
\end{tabular}
\end{table}

\begin{figure}[htb!]
    \centering
    \includegraphics[width=1\linewidth]{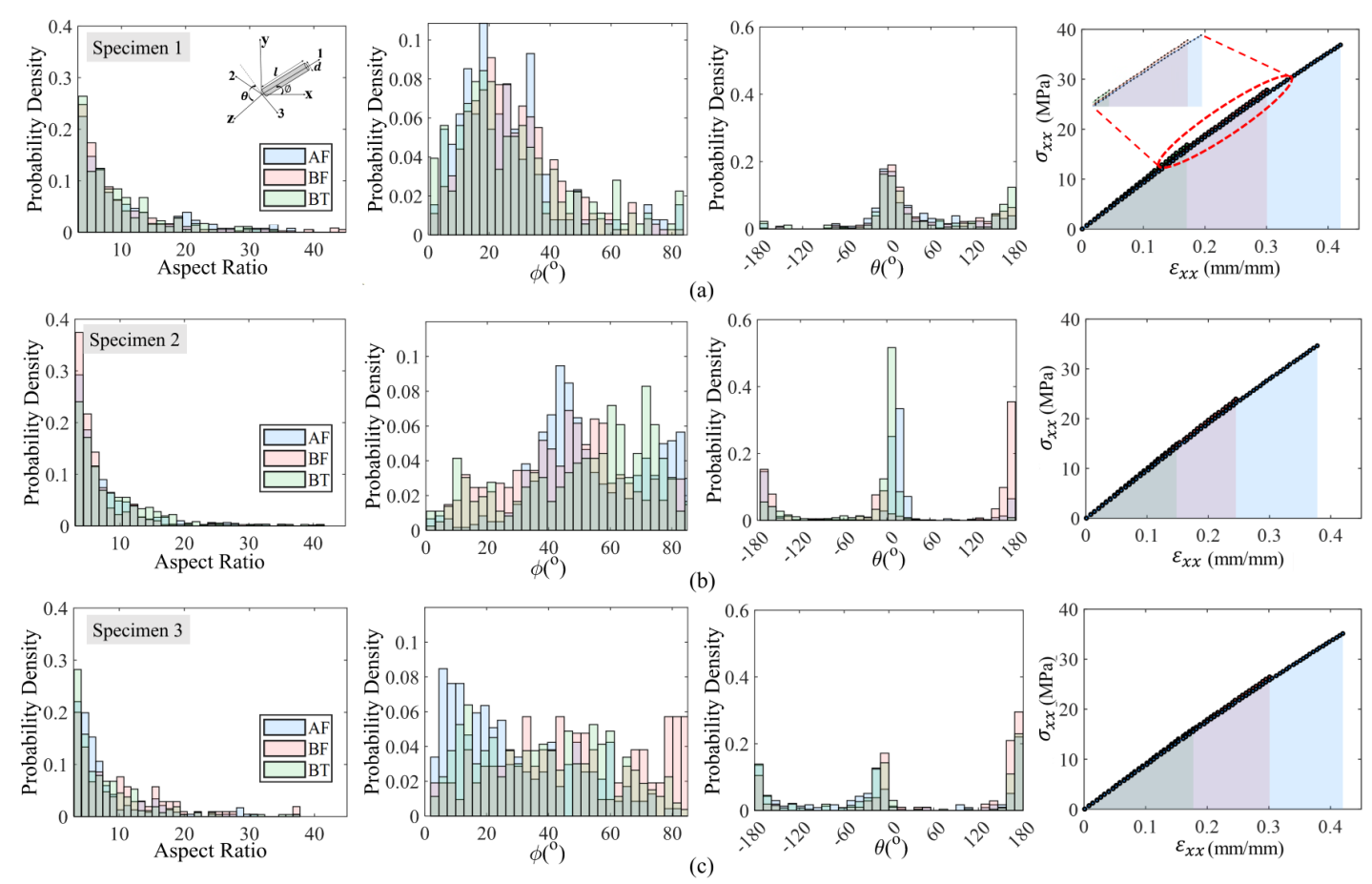}
    \caption{Statistical characterization of fiber morphology and its correlation with mechanical response across fatigue stages. \textbf{Left}: Histograms of measured fiber aspect ratios at each fatigue stage (BT, BF, AF), revealing quantifiable shifts in fiber length distribution under cyclic loading. \textbf{Center}: Histograms of the in-plane fiber orientation angle $\phi$ measured from the x-axis and angle $\theta$ in the y-z plane. \textbf{Right}: Experimentally measured axial stress-strain responses in fiber-dominated direction: before fatigue testing (BT), below-fatigue limit (BF), and above fatigue limit (AF). Rows correspond to Specimen~1 (top), Specimen~2 (middle), and Specimen~3 (bottom), which span three distinct initial porosity levels.}
    \label{fig:10}
\end{figure}

The extracted descriptors revealed a gradual yet systematic microstructural evolution under fatigue loading. Figure~\ref{fig:10} shows shifts in orientation distributions and aspect-ratio statistics, indicative of microstructural rearrangement and spatial reallocation of fiber-interaction neighborhoods under cyclic loading. Consistent with these observations, the porosity volume fraction increased progressively under cyclic loading for all three specimens (Table~\ref{tab:porosity_volume}). The presence of pores reduces the effective load-bearing capacity of the matrix and modifies the local stress state within the composite. The increase is particularly pronounced for specimens loaded above the fatigue limit, where porosity growth exceeded the initial (pre-fatigue) state by threefold. Although these changes were moderate at the ensemble level, their spatial localization significantly amplifies the progression of localized damage.

When correlated with the macroscale mechanical response (Figure~\ref{fig:10}), the progressive increase in porosity and localized damage accumulation corresponded to only marginal reductions in the longitudinal elastic stiffness and a subtle, early onset of axial nonlinear deformation. This highly stable axial behavior is consistent with the fiber-dominated nature of short-fiber architectures, where the longitudinal modulus is governed primarily by the high-stiffness reinforcement phase and remains relatively insensitive to low-to-moderate variations in matrix void content within the $1\%$--$6\%$ range~\cite{kumar2022anisotropic}. However, as fatigue-induced porosity grows, it severely disrupts matrix continuity and degrades matrix-mediated shear-lag load transfer between adjacent fibers. Consequently, while the axial modulus experiences minor degradation, this microstructural breakdown exerts a far more pronounced influence on the matrix- and interface-dominated mechanical properties, specifically the transverse normal and shear responses where voids act as localized stress concentrators~\cite{liu2022effect, NAIRN199763}. Importantly, this distinct sensitivity establishes that the severity of anisotropic stiffness loss is dictated by the local topology of the fiber network and inter-fiber matrix ligaments rather than a uniform volumetric loss of load-bearing material~\cite{Abhilash_2020, pathak_porosity_2025}.

Taken together, these experimental observations establish a clear, path-dependent microstructure--property relationship. The localized geometric heterogeneity within the material defines the initial fiber interaction topology; cyclic loading systematically modifies this topology through localized porosity growth and microstructural rearrangement, and the evolved topology ultimately governs the subsequent quasi-static mechanical response. These experimentally characterized features form the direct physical basis for the mesoscale interaction-based modeling framework utilized in the following section.

\subsection{Microstructure-Dependent Stiffness from Finite Element Homogenization}
\begin{figure}[htb!]
    \centering
    \includegraphics[width=1\linewidth]{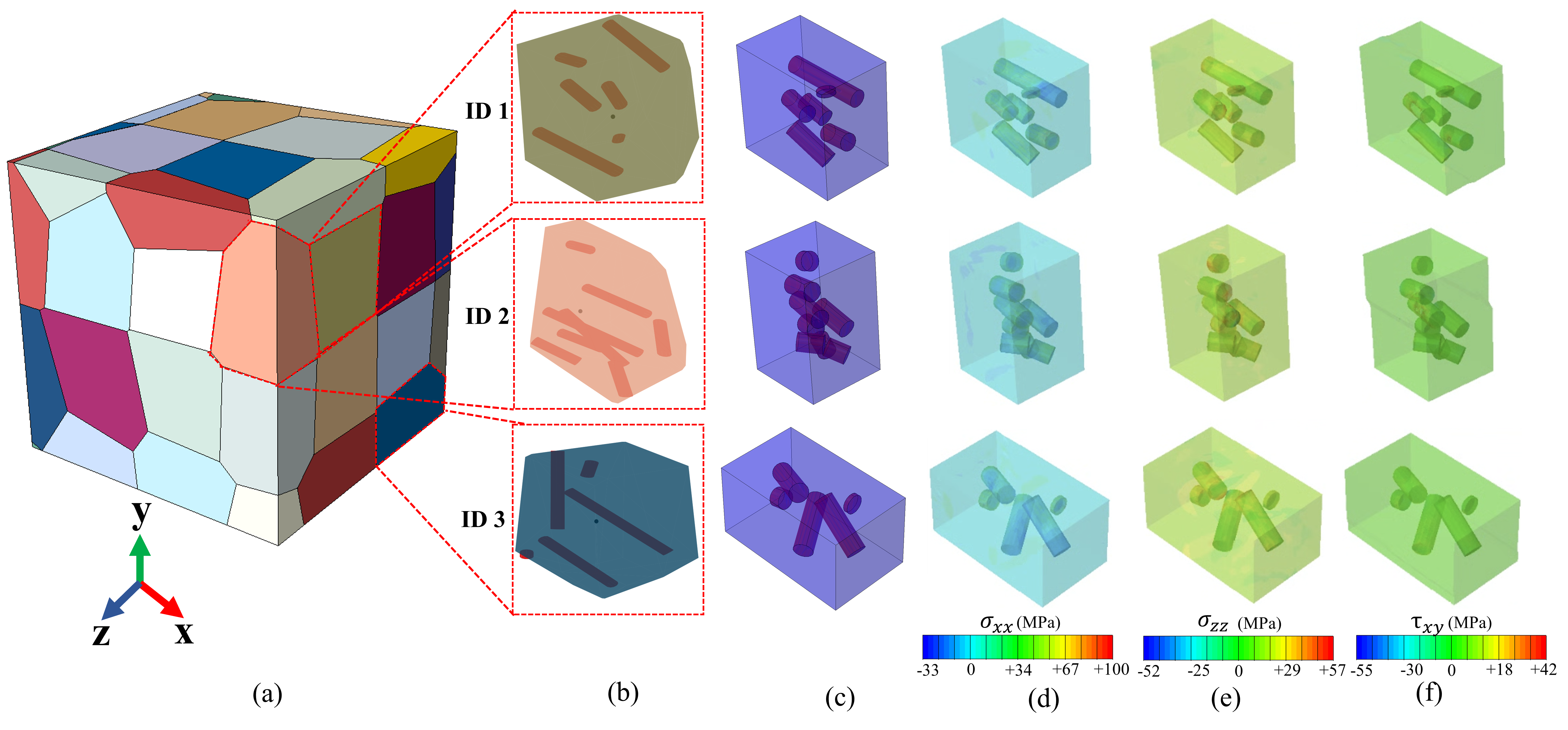}
    \caption{Discrete-cell multiscale modeling framework. (a) Mesoscale Voronoi tessellation of the reconstructed AM-CM microstructure. (b) Three representative polyhedral discrete cells (ID 1, ID 2, ID 3) depicting different fiber topologies. (c) Equivalent RUC FE models with explicitly resolved fiber geometries embedded in a homogenized porous matrix. (d–f) Representative local stress fields under: fiber-aligned normal loading in the x-direction (left), transverse loading in the z-direction (center), and axial shear in the x-y plane (right). Color scales are independently normalized per loading case to highlight intra-cell stress distribution patterns.}
    \label{fig:11}
\end{figure}

\begin{figure}[htb!]
    \centering
    \includegraphics[width=1\linewidth]{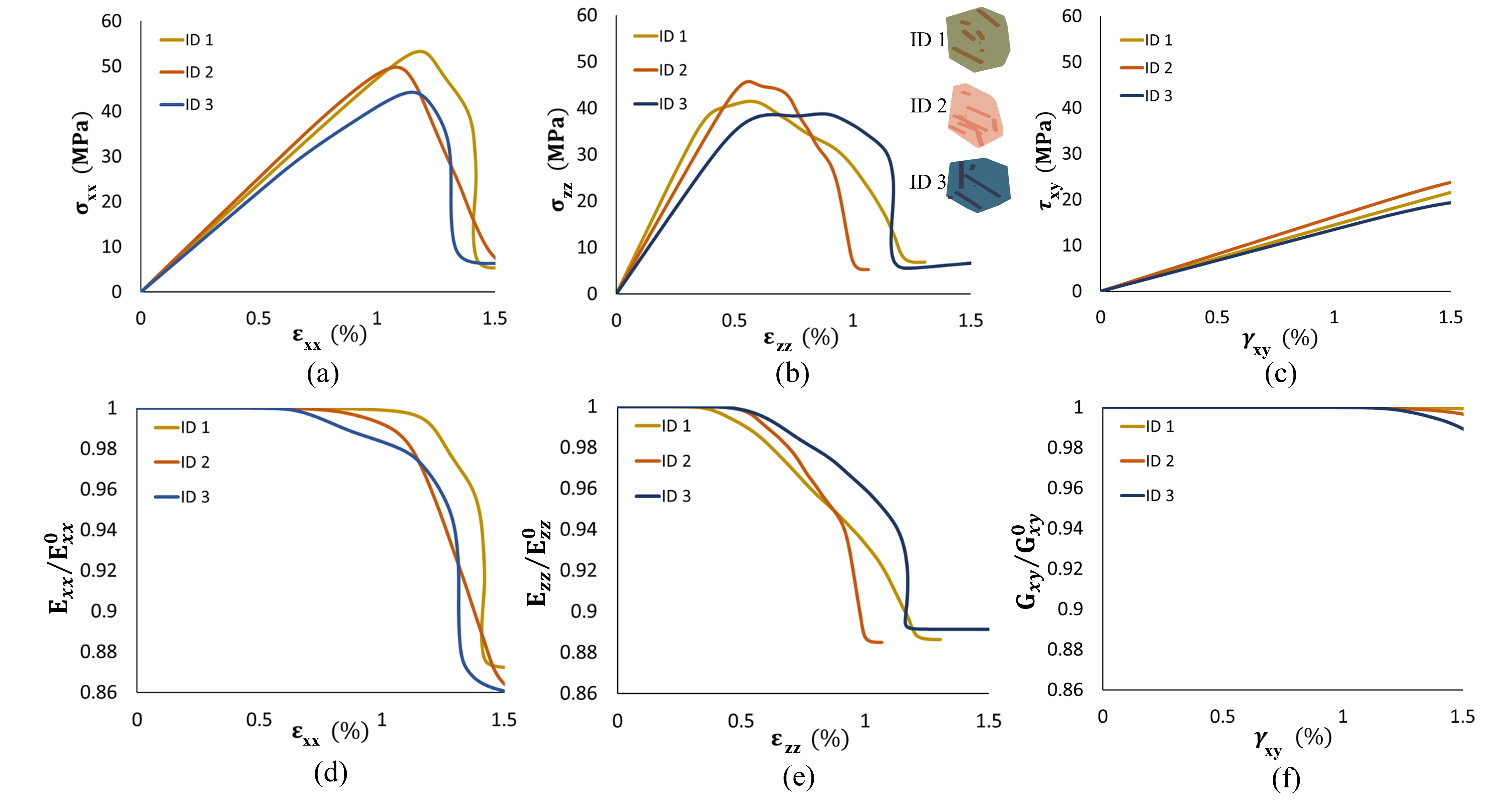}
    \caption{Nonlinear mechanical responses for the chosen discrete cells (ID 1, ID 2, ID 3). \textbf{Top row}: predicted stress–strain curves under (a) fiber-aligned normal loading ($\sigma_{xx}$–$\varepsilon_{xx}$), (b) transverse normal loading ($\sigma_{zz}$–$\varepsilon_{zz}$),  and (c) axial shear loading ($\tau_{xy}$–$\gamma_{xy}$). \textbf{Bottom row}: corresponding normalized tangent stiffness degradation expressed as (d) $E_{xx}/E_{xx}^{0}$, (e) $E_{zz}/E_{zz}^{0}$, and (f) $G_{xy}/G_{xy}^{0}$ as functions of applied strain.}
    \label{fig:12}
\end{figure}

Figure~\ref{fig:11} presents the local stress distributions for three discrete Voronoi cells subjected to identical far-field loading conditions. As shown in Figure~\ref{fig:11}a--c, these Voronoi cells were selected to represent contrasting microstructural topologies, varying significantly in their local fiber volume fractions, spatial alignments, clustering, and inter-fiber spacing. When subjected to simulated loading, these distinct microstructural descriptors produced markedly different mechanical responses, as reflected in the localized stress-field heterogeneity across panels Figure~\ref{fig:11}d--f and quantified by the macroscopic stress--strain trajectories in Figure~\ref{fig:12}.

Under axial loading along the x-axis ($\sigma_{xx}$), cells containing clusters of nearly aligned fibers (such as ID 1) exhibited pronounced stress localization along the fiber axes. This pattern indicates highly efficient load transfer via axial fiber stretching, yielding a correspondingly high effective longitudinal stiffness. Conversely, cells characterized by dispersed or obliquely oriented fibers (such as ID 3) exhibited more diffuse stress fields, in which a significantly larger fraction of the applied far-field load was carried by the surrounding thermoplastic matrix rather than the reinforcement phase. Because the continuous polymer matrix is substantially more compliant than the carbon fibers, these matrix-dominated cells exhibit an earlier onset of localized plastic deformation. The resulting variation in initial axial stiffness among the three representative cells examined here reaches approximately 8\%--12\%, highlighting the exceptional sensitivity of macroscale load-transfer efficiency to localized fiber topology.

Under in-plane shear loading ($\tau_{xy}$), the stress concentrations arose predominantly within the thin matrix ligaments separating adjacent fibers rather than within the fibers themselves. This shifts the structural burden, indicating that shear transfer across the microstructural domain was controlled primarily by matrix-mediated interactions rather than direct fiber reinforcement. Consequently, the axial shear stiffness was more strongly governed by local inter-fiber spacing and matrix continuity than by absolute fiber orientation---a mechanistic distinction with critical implications for the anisotropic damage evolution observed at later, highly cycled fatigue stages.

The cell-level stress--strain responses shown in Figure \ref{fig:12}a--c further illustrate the influence of local fiber topology on the resulting nonlinear constitutive behavior. Cells whose fibers were preferentially aligned with the primary loading axis exhibited both a higher initial elastic modulus and a more gradual transition to nonlinear softening. This observation is consistent with classical shear-lag theory, where the axial stress-transfer capacity and structural reinforcement efficiency scale directly with fiber orientation and aspect ratio. In contrast, cells with fewer load-aligned fibers exhibited lower initial stiffness and an accelerated onset of matrix-dominated nonlinearity.

The normalized tangent stiffness degradation curves in Figure \ref{fig:12}d--f (where $E_{ij}/E_{ij}^{0}$ denotes the ratio of instantaneous to initial undamaged stiffness) reveal three mechanistically significant trends. First, cells with well-connected fiber networks maintained their stiffness over a broader strain range, as load redistribution through multiple fiber--fiber interaction pathways remains operative even as matrix damage progresses. Second, cells characterized by sparse or poorly connected fiber networks exhibited earlier stiffness reduction, consistent with an accelerated transition from fiber- to matrix-dominated load-bearing. Third, axial shear stiffness degraded more gradually than normal stiffness, reflecting the comparatively stable matrix-controlled shear transfer mechanism.

These observations bear close resemblance to connectivity-driven stiffness transitions in disordered fiber networks, where the density and topology of load-bearing links dictate the onset and rate of stiffness loss \cite{Head_2003, torquato_random_2002}. In the present context, locally connected fiber clusters enable load redistribution via redundant pathways, thereby delaying macroscopic stiffness degradation, whereas sparsely connected or resin-rich cells offer limited redundancy and fail earlier via matrix plasticity and void growth. These results establish that local fiber-network connectivity---rather than volume fraction alone---governs the nonlinear evolution of stiffness in AM-CM SFTs, and that the developed graph-based surrogate modeling framework accurately captures this physical mechanism.

\subsection{Surrogate Prediction of Cell-Level Constitutive Behavior}
\label{Surrogate}
Direct nonlinear FE homogenization of every discrete cell is computationally prohibitive for mesoscale assemblies comprising hundreds to thousands of Voronoi cells. To enable efficient evaluation of cell-level constitutive behavior while preserving sensitivity to microstructural topology, the hybrid GNN-LSTM surrogate model described in Section \ref{GNN_LSTM} was trained on FE-generated stress--strain curves. The training dataset comprised stress--strain curves from 500 discrete Voronoi cells extracted from the $\mu$-CT microstructure, each evaluated under six independent loading cases (three normal and three shear), capturing a wide range of local fiber volume fractions, fiber orientations, and clustering configurations inherent to the AM-CM microstructure. The resulting dataset was partitioned using a stratified random shuffle (seed 42) into training (70\%), validation (15\%), and testing (15\%).

\begin{figure}[htb!]
    \centering
    \includegraphics[width=0.5\linewidth]{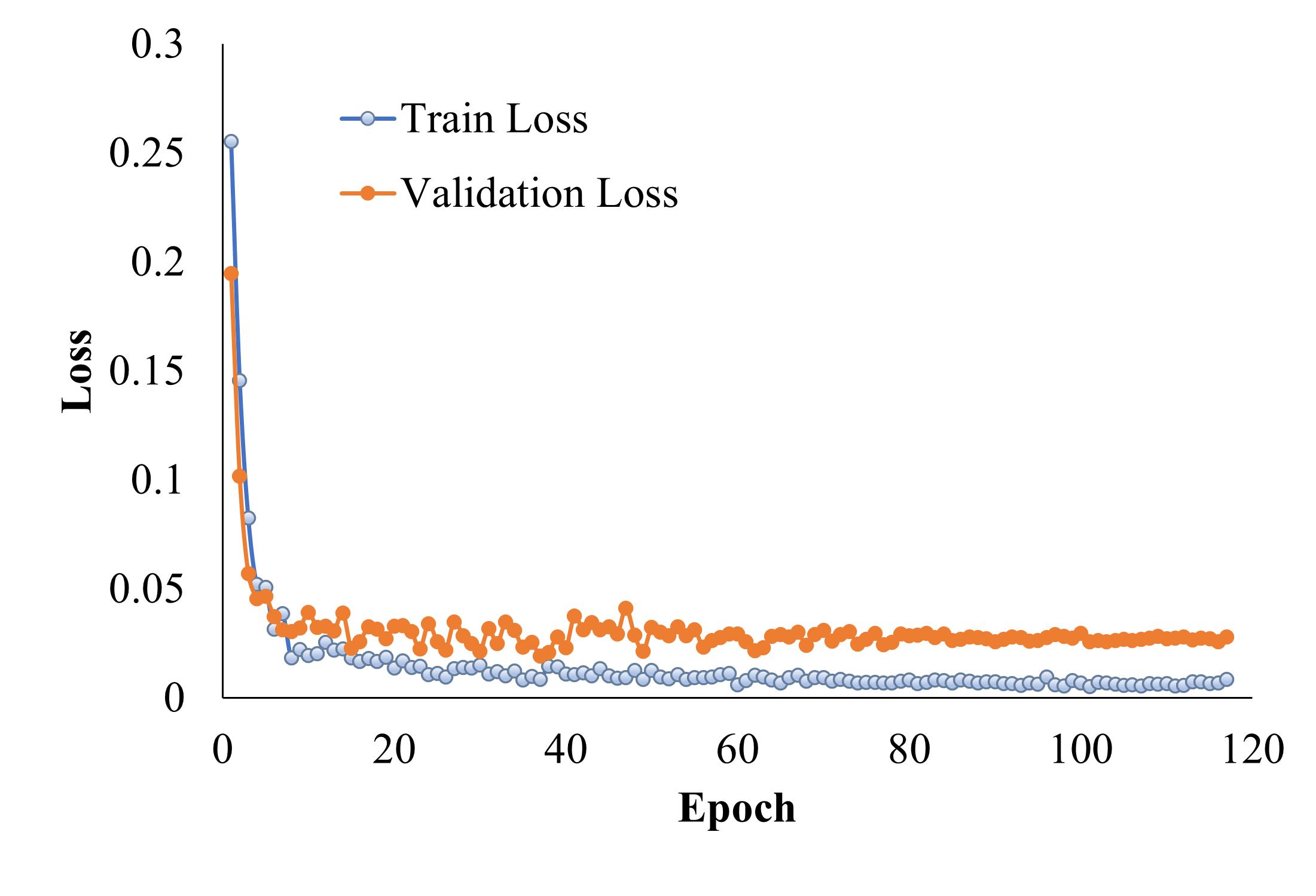}
    \caption{Training and validation loss histories demonstrating convergence of the GNN–LSTM surrogate for discrete-cell homogenization under normal loading in the fiber-dominated direction ($\sigma_{xx}$)}
    \label{fig:13_1}
\end{figure}

\textbf{\textit{Training convergence}}: The training history (Figure \ref{fig:13_1}) exhibited two characteristic phases. A rapid initial reduction in loss was associated with the surrogate learning the topology-dependent elastic stiffness signatures encoded in the GNN embedding, and the peak stress predictions from the auxiliary head and Random Forest regressor. This was followed by a more gradual convergence phase in which the LSTM captured the path-dependent plasticity and post-peak softening behavior governing the nonlinear damage regime. The strain-weighted Huber loss, which assigned up to $4\times$ higher weight near and beyond the stress peak, directed the majority of the learning capacity toward accurate representation of the mechanically critical nonlinear regime. The absence of divergence between training and validation loss throughout both phases confirmed stable generalization without overfitting, supporting the adequacy of the 500-cell dataset for the microstructural variability encountered. Early stopping was triggered at convergence, well within the 800-epoch maximum.

\begin{figure}[t]
    \centering
    \includegraphics[width=1\linewidth]{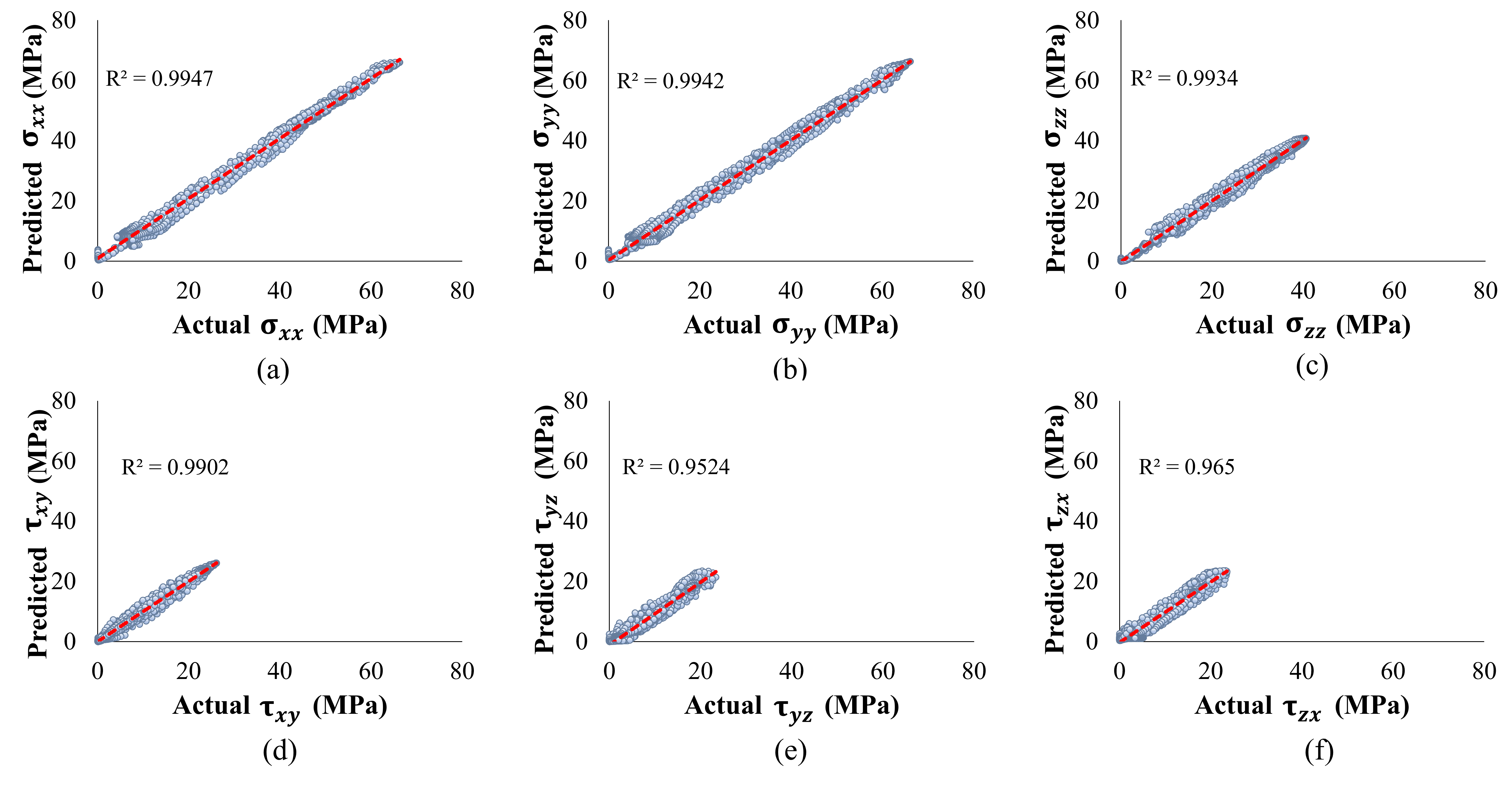}
    \caption{Parity plots comparing the high-fidelity FE homogenization benchmarks against the predictions generated by the hybrid GNN–LSTM surrogate framework. Individual coefficient of determination ($R^2$) values are reported across the six independent stress components: (a) fiber-dominated normal $\sigma_{xx}$, (b) transverse width normal $\sigma_{yy}$, (c) thickness normal $\sigma_{zz}$, (d) transverse shear $\tau_{xy}$, (e) axial shear $\tau_{yz}$, and (f) axial shear $\tau_{zx}$.}
    \label{fig:13_2}
\end{figure}

\textbf{\textit{Predictive accuracy}}:
A parity plot (Figure \ref{fig:13_2}a--f) demonstrated near-perfect agreement between surrogate-predicted and FE-computed stresses across the full stress range. This level of accuracy confirmed that the learned graph representation encoded physically meaningful microstructural information, capturing the topology-to-response mapping across the full diversity of fiber arrangements present in the $\mu$-CT microstructure---aligned fiber clusters, mixed-orientation neighborhoods, and resin-rich regions---rather than relying on globally averaged descriptors that would obscure topology-driven variability.

\begin{figure}[t]
    \centering
    \includegraphics[width=1\linewidth]{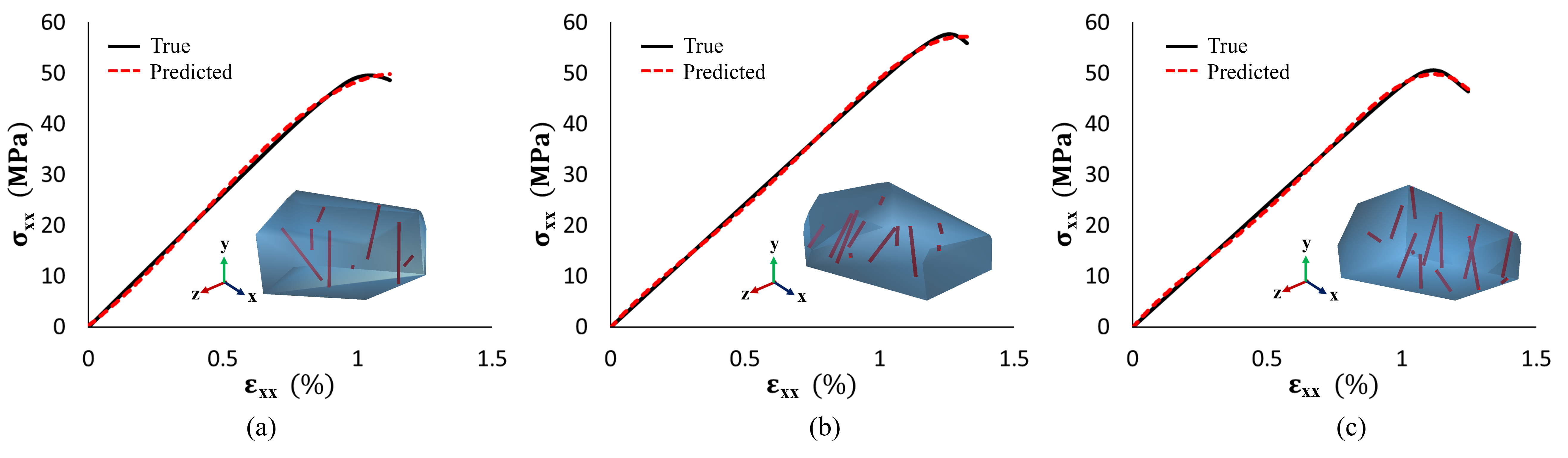}
    \caption{Representative comparison between nonlinear FE homogenization results (True) and GNN–LSTM surrogate predictions (Predicted) for three discrete Voronoi cells subjected to fiber-dominated normal loading. Each panel corresponds to a cell with a distinct local fiber topology: an aligned fiber cluster (a), a mixed-orientation neighborhood (b), and a resin-rich region (c).}
    \label{fig:14}
\end{figure}

Representative comparisons between FE simulations and surrogate predictions across a range of fiber topologies are presented in Figure \ref{fig:14}. The surrogate reproduced both the initial elastic stiffness and the post-yield nonlinear regime with high fidelity for cells characterized by aligned fiber clusters, mixed-orientation neighborhoods, and resin-rich regions, achieving $R^2 = 0.99$. Because all cells shared identical constituent material parameters, the variability in predicted response arose exclusively from differences in the local fiber topology encoded by the GNN---fiber count, spatial arrangement, orientation, distribution, and pairwise interaction strengths captured through the shear-lag edge weights. This confirmed that the surrogate learned a topology-sensitive constitutive mapping rather than a globally averaged response, and that the GNN embedding $z_{\Omega_c}$ successfully encoded the mechanically distinguishing features of each cell's fiber network.

\textbf{\textit{Computational efficiency}}:
The trained surrogate predicted the complete normalized stress--strain trajectory of an individual cell in under one second on a single CPU, with subsequent de-normalization via the Random Forest peak-stress prediction adding negligible computational cost. The total per-cell prediction time therefore remained under one second, compared with several minutes required for the equivalent nonlinear FE homogenization running on 5 CPUs in Abaqus 2022. This reduction in computational cost, by more than two orders of magnitude, enabled efficient evaluation of hundreds to thousands of discrete cells within a mesoscale Voronoi assembly and rendered the proposed multiscale framework tractable for fatigue-stage-resolved analyses across multiple specimens and damage states. Furthermore, unlike FE homogenization where cells are coupled through boundary conditions, surrogate evaluation of individual Voronoi cells is completely independent once the macroscopic strain is localized to each cell, meaning all cells could in principle be evaluated simultaneously in a parallelized implementation, with the theoretical speedup scaling linearly with the number of available processors---an advantage that was not exploited in the present single-CPU implementation but represents a promising avenue for near-real-time coupon-scale predictions in future HPC deployments.

\subsection{Mesoscale Homogenization and Coupon-Scale Validation}
The ultimate purpose of the proposed multiscale framework is to establish a quantitative, predictive link between experimentally observed microstructural degradation and macroscopic mechanical response. To evaluate this capability, mesoscale Voronoi assemblies were constructed directly from $\mu$-CT reconstructions of fatigue-tested specimens. Three specimens, each representing a distinct interrupted fatigue stage (Section~\ref{sec:manufacturing}), were selected to capture progressive damage states ranging from nominally undamaged to near-failure.

For each specimen and fatigue stage, the fiber centerline coordinates and cell-level volume fractions extracted from the $\mu$-CT reconstruction were passed directly to the GNN-LSTM surrogate to predict cell-level constitutive responses, which were assembled into the coupon-scale response. No constitutive parameters were recalibrated at the coupon scale between fatigue stages; all observed changes in the predicted macroscopic response arose solely from the microstructure-informed constitutive update propagated through the LSTM component of the surrogate, in which the cell parameters are uniformly applied across all time points.

At each fatigue stage, the homogenization procedure incorporated $\mu$-CT-measured porosity through the two-step porous-matrix model described in Section~\ref{2-step homogenization}, which reduced the effective matrix stiffness assigned to all Voronoi cells. The GNN-LSTM then predicted the normalized stress trajectory for each cell based on its $\mu$-CT fiber topology, and the Random Forest predicted the corresponding peak stress amplitude from the GNN embedding and local volume fraction. As fatigue damage advanced, increasing porosity reduced the porous-matrix stiffness, loss of fiber connectivity, and cluster dispersion were directly captured in lower GNN edge weights and reduced topology embeddings, and the Random Forest consequently predicted lower peak stresses for damaged cells, collectively degrading the assembled coupon response in a manner driven entirely by the evolving $\mu$-CT microstructure rather than by any manually adjusted parameter.

\begin{figure}[htp!]
    \centering
\includegraphics[width=1\linewidth]{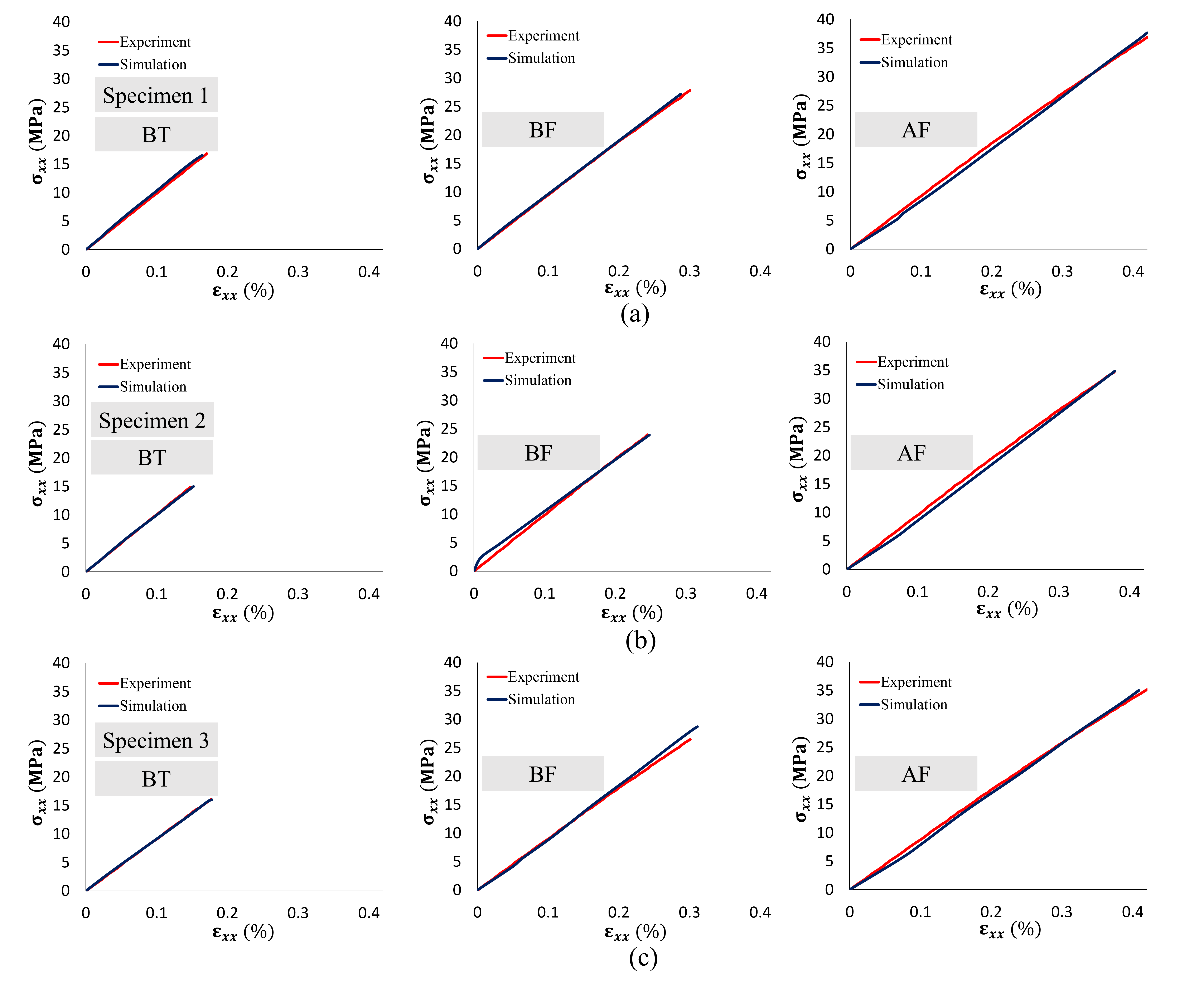}
    \caption{Comparison of experimentally measured and multiscale-predicted stress--strain responses obtained from the hybrid FE-surrogate framework, in which GNN-LSTM surrogate predictions of discrete Voronoi cell constitutive responses are assembled through mesoscale periodic boundary conditions and volume-weighted averaging to predict the coupon-scale response, for all three specimens at each fatigue stage (BT: before test, BF: below fatigue limit, AF: above fatigue limit). Top (a), middle (b), and bottom (c) panels correspond to Specimen 1, Specimen 2, and Specimen 3, respectively.}
\label{fig:15}
\end{figure}

As illustrated in Appendix~\ref{app:A3}, the full six-mode constitutive response is first obtained for each discrete Voronoi cell---either through FE homogenization or the GNN-LSTM surrogate---and these cell-level responses are then assembled through volume-weighted averaging at the mesoscale Voronoi assembly level to produce the final coupon-scale response. Figure~\ref{fig:15} shows the longitudinal mesoscale stress--strain response comparison with experimental monotonic tensile measurements. With increasing fatigue damage, both experiments and simulations showed a systematic reduction in elastic modulus and an earlier onset of nonlinear deformation. These trends were reproduced by the multiscale model without any coupon-scale recalibration, confirming that the observed macroscopic degradation was captured entirely through the microstructure-informed constitutive update propagated from $\mu$-CT observations through the surrogate.

The model reproduced the progressive softening of the stress--strain response across the three fatigue conditions. Quantitatively, predicted elastic moduli agreed with experimental measurements within 5--7\% across all three specimens and fatigue stages, confirming that the surrogate captured the correct sensitivity to microstructural degradation across the full range of damage studied, as shown in Figure~\ref{fig:15}.
The discrepancies between simulation and experiment were attributed primarily to physical mechanisms not represented in the current constitutive framework, rather than to surrogate approximation error. The matrix model captures elastoplastic damage and porosity-induced stiffness reduction, but does not explicitly resolve fiber-matrix interfacial debonding or discrete matrix cracking. Both become increasingly active at advanced stages of fatigue, causing further loss of stiffness in highly damaged regions. The surrogate itself contributed negligibly to this error, as confirmed by comparing coupon-scale predictions from the GNN-LSTM surrogate with those obtained from direct FE homogenization of the same Voronoi assemblies. These agree closely across all fatigue stages, isolating the source of the experiment-simulation gap to the constitutive model rather than to the surrogate model.

Taken together, these results demonstrated that the proposed FE-Voronoi-GNN framework successfully propagated microstructural information across three length scales---from the explicitly resolved fiber level through the topology-sensitive surrogate to the macroscopic coupon response---using only $\mu$-CT fiber coordinates and volume fractions as inputs, with no specialized characterization beyond standard reconstruction. By directly linking manufacturing-induced and fatigue-driven microstructural evolution to measurable changes in mechanical performance, the framework provided a physically grounded and computationally efficient basis for assessing the structural condition and remaining fatigue life of AM-CM short-fiber thermoplastic composite components.

It is worth noting that while a mean-field RVE approach using average microstructural descriptors may yield comparable elastic modulus predictions at the coupon scale, the present framework offers capabilities that such an approach fundamentally cannot provide: (i)~spatially resolved identification of mechanically weak cells governed by local fiber clustering and sparse connectivity, (ii)~direct propagation of fatigue-stage-specific microstructural evolution without constitutive recalibration, and (iii)~capture of topology-driven variability in nonlinear damage evolution that is entirely lost under mean-field averaging.

\section{Conclusions and Future Work}
This study established a microstructure-resolved multiscale surrogate modeling framework to predict the nonlinear mechanical response of SFT composites manufactured via AM-CM. By combining interaction-topology-based discrete homogenization with a physics-informed GNN-LSTM surrogate, the framework provided an efficient, physically interpretable pathway for linking local fiber topology directly to macroscopic structural performance. 

The key findings are summarized below.

\begin{itemize}

\item \textbf{Microstructure-controlled mechanics in discrete cells:}  
High-resolution $\mu$-CT reconstructions demonstrate that the AM-CM microstructure can be decomposed into localized Voronoi cells representing neighborhoods of interacting fibers. Even when global fiber volume fraction and constituent properties remain unchanged, these cells exhibit substantial variability in stiffness and nonlinear response arising solely from differences in fiber clustering, alignment, and spacing. The mechanical behavior of SFTs is therefore governed not only by volume fraction but also by the topology of local fiber interaction networks that control load-transfer pathways between fibers and matrix. Nonlinear FE homogenization showed that cells containing aligned or highly connected fiber clusters sustained load primarily through axial fiber stretching, yielding higher stiffness and delayed degradation, whereas cells with dispersed or sparsely connected configurations relied more on matrix-mediated load transfer, producing earlier nonlinear deformation and accelerated stiffness reduction.

\item \textbf{Physics-informed surrogate modeling:}  
A hybrid GNN-LSTM architecture was developed and trained to approximate the nonlinear constitutive response of discrete cells. The surrogate required 3D fiber centerline endpoints and cell-level volume fractions as inputs, both of which are available directly from standard $\mu$-CT reconstruction. By representing the fiber interaction network as a graph with physics-informed edge weights that encode both axial shear-lag coupling and the orientation-independent transverse stress-concentration mechanism, as confirmed by parametric two-fiber FEA, the GNN captured topology-dependent stiffness.

The LSTM learned the path-dependent normalized constitutive behavior; a skip connection from the GNN embedding to the output network ensured topology-encoded amplitude information directly modulated stress predictions at every loading increment. This two-stage prediction pipeline reproduced FE-predicted stress--strain responses with $R^2\approx0.98$ on de-normalized physical stress trajectories while reducing per-cell evaluation time by more than two orders of magnitude, enabling near-real-time prediction over hundreds to thousands of Voronoi cells.

\item \textbf{Multiscale propagation of microstructural degradation.}  
When integrated within mesoscale Voronoi assemblies derived directly from $\mu$-CT reconstructions at each fatigue stage, the surrogate-enabled framework accurately predicted coupon-level tensile responses across multiple damage states and three specimens without recalibrating any constitutive parameters. Predicted elastic moduli agreed with experiment within 5--7\%, and area under the stress--strain curve within 10\%, confirming that microstructure-driven degradation mechanisms were correctly propagated across scales through the surrogate alone, driven entirely by the evolving $\mu$-CT microstructure.

\end{itemize}

Collectively, these results established that the mechanical response of AM-CM SFTs was fundamentally governed by the topology of local fiber interaction networks rather than by global microstructural averages alone. The proposed FE--Voronoi--GNN framework established a physically grounded bridge between microstructural evolution and macroscopic mechanical performance, deployable on any $\mu$-CT dataset from which fiber centerlines and volume fractions could be extracted, with no specialized inputs beyond standard reconstruction.

Beyond the specific CF-ABS system studied, the framework provides a scalable foundation for microstructure-aware digital twins of discontinuous fiber composites. The ability to predict constitutive response rapidly and directly from minimal microstructural inputs opens opportunities for accelerated materials design, process optimization, and structural performance assessment in additively manufactured composite systems.

Future work will focus on:
(i)~incorporating explicit fiber-matrix interfacial debonding mechanisms to capture intra-cell failure modes not currently represented;
(ii)~embedding thermodynamic admissibility constraints directly within the GNN-LSTM architecture to enforce energetic consistency for more generalized global loading histories, including cyclic fatigue loading and non-proportional multiaxial loading paths at the coupon scale, beyond the monotonic quasi-static regime studied here; and
(iii)~extending the framework to other thermoplastic composite systems and processing routes, including injection molding and multi-material additive manufacturing.

\section{Acknowledgments}
This research was supported by the NASA 80NSSC24M0154 Grant. This research was also supported in part by the U.S. Department of Energy Office of Energy Efficiency and Renewable Energy (EERE), Advanced Materials and Manufacturing Technologies Office (AMMTO).

The authors would like to express their sincere gratitude to Brandon L. Hearley at the NASA Glenn Research Center for his valuable review, insights, and assistance in the preparation and writing of this manuscript.

The authors utilized AI-based tools to assist with proofreading and refining the manuscript's clarity. Specifically, Claude and Grammarly were employed to improve grammar, sentence flow, and readability. No generative AI tools were used for data analysis, interpretation of results, or drawing scientific conclusions. The responsibility for the content and final approval of the manuscript rests entirely with the authors.
\newpage
\begin{singlespace}
\bibliographystyle{elsarticle-num}
\bibliography{myreference} 

@article{gunst_leveraging_2025,
    title = {Leveraging Machine Learning to Quantify the Effect of Porosity on Mode I Delamination in Ceramic Matrix Composites},
    author = {Gunst, Joseph and Vaghefi, Ehsan and Mirkoohi, Elham and Deshpande, Gopikrishna and Prorok, Barton and Gururaja, Suhasini},
    journal = {Composite Structures},
    volume = {372},
    pages = {119542},
    year = {2025},
    doi = {10.1016/j.compstruct.2025.119542},
}

@article{Pothnis_2019,
title = {Development and characterization of electric field directed preferentially aligned CNT nanocomposites},
journal = {Mechanics of Advanced Materials and Structures},
volume = {26},
pages = {35--41},
year = {2019},
doi = {10.1080/15376494.2018.1534165},
author = {Pothnis, J.R. and Gururaja, S. and Kalyanasundaram, D.}
}

@article{Kammoun_2015,
title = {Micromechanical modeling of the progressive failure in short glass–fiber reinforced thermoplastics – First Pseudo-Grain Damage model},
journal = {Composites Part A: Applied Science and Manufacturing},
volume = {73},
pages = {166-175},
year = {2015},
issn = {1359-835X},
doi = {10.1016/j.compositesa.2015.02.017},
author = {Kammoun, S. and Doghri, I. and Brassart, L. and Delannay, L.}
}

@article{Breuer_2021,
title = {Prediction of Short Fiber Composite Properties by an Artificial Neural Network Trained on an RVE Database},
journal = {Fibers},
volume = {9},
pages = {8},
year = {2021},
issn = {2079-6439},
doi = {10.3390/fib9020008},
author = {Breuer, Kevin and Stommel, Markus}
}

@article{Hao_2022,
author = {Xu, Hao and Kuczynska, Marta and Schafet, Natalja and Welschinger, Fabian and Hohe, Jörg},
title ={FE-based damage modeling approach for short fiber reinforced thermoplastics under quasi-static load coupling anisotropic viscoplasticity and matrix degradation},
journal = {Journal of Composite Materials},
volume = {56},
pages = {3113-3125},
year = {2022},
doi = {10.1177/00219983221109327},
}

@article{Arabatti_2016,
author = {Arabatti, Trupti and Parambil, Nithin K. and Gururaja, Suhasini},
title ={Micromechanical Modeling of Damage Development in Polymer Composites},
journal = {Advanced Composites Letters},
volume = {25},
pages = {096369351602500301},
year = {2016},
doi = {10.1177/096369351602500301},
}

@article{LI_2004,
title = {Unit cells for micromechanical analyses of particle-reinforced composites},
journal = {Mechanics of Materials},
volume = {36},
pages = {543-572},
year = {2004},
issn = {0167-6636},
doi = {10.1016/S0167-6636(03)00062-0},
author = {Shuguang, Li and Anchana, Wongsto},
}

@article{chen_deep_2021,
    title = {Deep Long Short-Term Memory Neural Network for Accelerated Elastoplastic Analysis of Heterogeneous Materials: An Integrated Data-Driven Surrogate Approach},
    author = {Chen, Qiang and Jia, Ruijian and Pang, Shanmin},
    journal = {Composite Structures},
    volume = {264},
    pages = {113688},
    year = {2021},
    doi = {10.1016/j.compstruct.2021.113688},
}

@article{jiang_design_2022,
    title = {Design of Short Fiber-Reinforced Thermoplastic Composites: A Review},
    author = {Jiang, Lijuan and Zhou, Yinzhi and Jin, Fengnian},
    journal = {Polymer Composites},
    volume = {43},
    pages = {4835--4847},
    year = {2022},
    doi = {10.1002/pc.26817},
}

@article{zhao_high-generalizability_2023,
    title = {A High-Generalizability Machine Learning Framework for Analyzing the Homogenized Properties of Short Fiber-Reinforced Polymer Composites},
    author = {Zhao, Y. and Chen, Z. and Jian, X.},
    journal = {Polymers},
    volume = {15},
    pages = {3962},
    year = {2023},
    doi = {10.3390/polym15193962},
}

@article{cassola_machine_2022,
    title = {Machine Learning for Polymer Composites Process Simulation – a Review},
    author = {Cassola, Stefano and Duhovic, Miro and Schmidt, Tim and May, David},
    journal = {Composites Part B: Engineering},
    volume = {246},
    pages = {110208},
    year = {2022},
    doi = {10.1016/j.compositesb.2022.110208},
}

@article{sharma_review_2025,
    title = {Review of Machine Learning Approaches for Predicting Mechanical Behavior of Composite Materials},
    author = {Sharma, Harshit and Arora, Gaurav and Singh, Manoj Kumar and Ayyappan, Vinod and Bhowmik, Papiya and Rangappa, Sanjay Mavinkere and Siengchin, Suchart},
    journal = {Discover Applied Sciences},
    volume = {7},
    pages = {1238},
    year = {2025},
    doi = {10.1007/s42452-025-07616-8},
}

@article{ferdousi_deep_2025,
    title = {A Deep Learning and Finite Element Approach for Exploration of Inverse Structure–Property Designs of Lightweight Hybrid Composites},
    author = {Ferdousi, Sanjida and Demchuk, Zoriana and Choi, Wonbong and Advincula, Rigoberto C. and Jiang, Yijie},
    journal = {Composite Structures},
    volume = {365},
    pages = {119179},
    year = {2025},
    doi = {10.1016/j.compstruct.2025.119179},
}

@article{kibrete_ai_2023,
    title = {Artificial Intelligence in Predicting Mechanical Properties of Composite Materials},
    author = {Kibrete, F. and Trzepieciński, T. and Gebremedhen, H. S. and Woldemichael, D. E.},
    journal = {Journal of Composites Science},
    volume = {7},
    pages = {364},
    year = {2023},
    doi = {10.3390/jcs7090364},
}

@article{chen_machine_2019,
    title = {Machine Learning for Composite Materials},
    author = {Chen, Chun-Teh and Gu, Grace X.},
    journal = {MRS Communications},
    volume = {9},
    pages = {556--566},
    year = {2019},
    doi = {10.1557/mrc.2019.32},
}

@article{kobler_fiber_2018,
	title = {Fiber orientation interpolation for the multiscale analysis of short fiber reinforced composite parts},
	volume = {61},
	issn = {0178-7675, 1432-0924},
	doi = {10.1007/s00466-017-1478-0},
	language = {en},
	journal = {Computational Mechanics},
	author = {Köbler, Jonathan and Schneider, Matti and Ospald, Felix and Andrä, Heiko and Müller, Ralf},
	year = {2018},
	pages = {729--750},
}

@article{eyri_modeling_2025,
	title = {Modeling short fiber reinforced polymer matrix composite materials using material designer},
	volume = {46},
	issn = {0272-8397, 1548-0569},
	doi = {10.1002/pc.29625},
	language = {en},
	journal = {Polymer Composites},
	author = {Eyri, Busra and Gul, Okan and Karsli, N. Gamze and Yilmaz, Taner},
	year = {2025},
	pages = {10350--10360},
}

@article{hamza_physics-constrained_2025,
	title = {Physics-constrained machine learning surrogate model for time-dependent behavior of ceramic matrix composites},
	volume = {307},
	issn = {1359-8368},
	doi = {10.1016/j.compositesb.2025.112825},
	journal = {Composites Part B: Engineering},
	author = {Hamza, M. H. and Borkowski, L. and Chattopadhyay, A.},
	year = {2025},
	pages = {112825},
}

@article{mentges_micromechanical_2023,
	title = {Micromechanical modelling of short fibre composites considering fibre length distributions},
	volume = {264},
	issn = {1359-8368},
	doi = {10.1016/j.compositesb.2023.110868},
	journal = {Composites Part B: Engineering},
	author = {Mentges, N. and Çelik, H. and Hopmann, C. and Fagerström, M. and Mirkhalaf, S. M.},
	year = {2023},
	pages = {110868},
}

@article{schraa_characterisation_2025,
	title = {Characterisation and modelling of the fibre-matrix interface of short fibre reinforced thermoplastics using the push-out technique},
	volume = {297},
	issn = {1359-8368},
	doi = {10.1016/j.compositesb.2025.112317},
	journal = {Composites Part B: Engineering},
	author = {Schraa, Lucas and Rodricks, Carol and Kalinka, Gerhard and Roetsch, Karl and Scheffler, Christina and Sambale, Anna and Uhlig, Kai and Stommel, Markus and Trappe, Volker},
	year = {2025},
	pages = {112317},
}

@article{wu_micro-mechanics_2021,
	title = {Micro-mechanics and data-driven based reduced order models for multi-scale analyses of woven composites},
	volume = {270},
	issn = {0263-8223},
	doi = {10.1016/j.compstruct.2021.114058},
	journal = {Composite Structures},
	author = {Wu, Ling and Adam, Laurent and Noels, Ludovic},
	year = {2021},
	pages = {114058},
}

@article{estefani_numerical_2024,
	title = {Numerical multiscale analysis of {3D} printed short fiber composites parts: {Filament} anisotropy and toolpath effects},
	volume = {6},
	issn = {2577-8196, 2577-8196},
	shorttitle = {Numerical multiscale analysis of {3D} printed short fiber composites parts},
	doi = {10.1002/eng2.12799},
	language = {en},
	journal = {Engineering Reports},
	author = {Estefani, Alejandro and Távara, Luis},
	year = {2024},
	pages = {12799},
}

@article{uhlig_meso-scaled_2016,
	title = {Meso-scaled finite element analysis of fiber reinforced plastics made by {Tailored} {Fiber} {Placement}},
	volume = {143},
	issn = {0263-8223},
	doi = {10.1016/j.compstruct.2016.01.049},
	journal = {Composite Structures},
	author = {Uhlig, K. and Tosch, M. and Bittrich, L. and Leipprand, A. and Dey, S. and Spickenheuer, A. and Heinrich, G.},
	year = {2016},
	pages = {53--62},
}

@online{kaiser_extended_2012,
	title = {An extended mean-field homogenization model to predict the strength of short-fibre polymer composites},
	volume = {32},
	journal = {Technische Mechanik},
	author = {Kaiser, Jan Martin and Stommel, Markus},
	year = {2012},
	pages = {307--320},
  howpublished = {Accessed: Dec. 29, 2025. [Online]. Available: \url{https://journals.ub.ovgu.de/index.php/techmech/article/view/725},
}
}

@article{li_mean-field_2025,
	title = {A mean-field homogenization model for fiber reinforced composite materials in large deformation with plasticity},
	volume = {310},
	issn = {0020-7683},
	doi = {10.1016/j.ijsolstr.2024.113200},
	journal = {International Journal of Solids and Structures},
	author = {L. Anqi and J.J.C. Remmers and A.W.J. Van Dommelen and J.T. Massart and G.D.M. Geers},
	year = {2025},
	pages = {113200},
}

@article{muller_homogenization_2015,
	title = {Homogenization of linear elastic properties of short-fiber reinforced composites – {A} comparison of mean field and voxel-based methods},
	volume = {67-68},
	issn = {0020-7683},
	doi = {10.1016/j.ijsolstr.2015.02.030},
	journal = {International Journal of Solids and Structures},
	author = {Müller, Viktor and Kabel, Matthias and Andrä, Heiko and Böhlke, Thomas},
	year = {2015},
	pages = {56--70},
}

@article{fu_post-mortem_2005,
	title = {On the post-mortem fracture surface morphology of short fiber reinforced thermoplastics},
	volume = {36},
	issn = {1359-835X},
	doi = {10.1016/j.compositesa.2004.11.005},
	journal = {Composites Part A: Applied Science and Manufacturing},
	author = {Fu, S. Y. and Lauke, B. and Zhang, Y. H. and Mai, Y.-W.},
	year = {2005},
	pages = {987--994},
}

@article{Mortazavian_2015,
title = {Fatigue behavior and modeling of short fiber reinforced polymer composites: A literature review},
journal = {International Journal of Fatigue},
volume = {70},
pages = {297-321},
year = {2015},
issn = {0142-1123},
doi = {10.1016/j.ijfatigue.2014.10.005},
author = {Seyyedvahid Mortazavian and Ali Fatemi},
}

@article{Tamboura_2020,
title = {Damage and fatigue life prediction of short fiber reinforced composites submitted to variable temperature loading: Application to Sheet Molding Compound composites},
journal = {International Journal of Fatigue},
volume = {138},
pages = {105676},
year = {2020},
issn = {0142-1123},
doi = {10.1016/j.ijfatigue.2020.105676},
author = {S. Tamboura and M.A. Laribi and J. Fitoussi and M. Shirinbayan and R. Tie Bi and A. Tcharkhtchi and H. Ben Dali},
}

@article{Vipin_2025,
title = {Genesis of a novel high-rate composite manufacturing process using large-scale additive manufacturing – compression molding (AM-CM) system: Possibilities and limitations∗},
journal = {Composites Part B: Engineering},
volume = {309},
pages = {113101},
year = {2026},
issn = {1359-8368},
doi = {10.1016/j.compositesb.2025.113101},
author = {Vipin Kumar and Ahmed Arabi Hassen and Vlastimil Kunc and David Nuttall and Anil Kircaliali and Segun Isaac Talabi and Jay Reynolds and Joshua Vaughan and Paritosh Mhatre and Tyler Smith and Berin Šeta and Jon Spangenberg and Craig Blue},
}

@article{rezaei_development_2008,
	title = {Development of {Short}-{Carbon}-{Fiber}-{Reinforced} {Polypropylene} {Composite} for {Car} {Bonnet}},
	volume = {47},
	issn = {0360-2559, 1525-6111},
	doi = {10.1080/03602550801897323},
	language = {en},
	journal = {Polymer-Plastics Technology and Engineering},
	author = {Rezaei, F. and Yunus, R. and Ibrahim, N. A. and Mahdi, E. S.},
	year = {2008},
	pages = {351--357},
}

@article{donal_1997,
	title = {Short Fiber Orientation and Its Effects on the Properties of Thermoplastic Composite Materials.},
	volume = {8},
	doi = {10.1080/03602557708545033},
	language = {en},
	journal = {Polymer-Plastics Technology and Engineering},
	author = {Donal, M.},
	year = {1997},
	pages = {101--54},
}

@article{gandhi_method_2016,
	title = {Method to measure orientation of discontinuous fiber embedded in the polymer matrix from computerized tomography scan data},
	volume = {29},
	issn = {0892-7057, 1530-7980},
	doi = {10.1177/0892705715584411},
	language = {en},
	journal = {Journal of Thermoplastic Composite Materials},
	author = {Gandhi, Umesh and Sebastian, De Boodt and Kunc, Vlastimil and Song, YuYang},
	year = {2016},
	pages = {1696--1709},
}

@article{tekinalp_highly_2014,
	title = {Highly oriented carbon fiber–polymer composites via additive manufacturing},
	volume = {105},
	issn = {0266-3538},
	doi = {10.1016/j.compscitech.2014.10.009},
	journal = {Composites Science and Technology},
	author = {Tekinalp, Halil L. and Kunc, Vlastimil and Velez-Garcia, Gregorio M. and Duty, Chad E. and Love, Lonnie J. and Naskar, Amit K. and Blue, Craig A. and Ozcan, Soydan},
	year = {2014},
	pages = {144--150},
}

@article{parandoush_review_2017,
	title = {A review on additive manufacturing of polymer-fiber composites},
	volume = {182},
	issn = {0263-8223},
	doi = {10.1016/j.compstruct.2017.08.088},
	journal = {Composite Structures},
	author = {Parandoush, Pedram and Lin, Dong},
	year = {2017},
	pages = {36--53},
}

@incollection{advani_3_2012,
title = {3 - Compression molding in polymer matrix composites},
booktitle = {Manufacturing Techniques for Polymer Matrix Composites (PMCs)},
publisher = {Woodhead Publishing},
pages = {47-94},
year = {2012},
series = {Woodhead Publishing Series in Composites Science and Engineering},
isbn = {978-0-85709-067-6},
doi = {10.1533/9780857096258.1.47},
author = {C.H. Park and W.I. Lee},
}

@article{dai_graph_2021,
	title = {Graph neural networks for an accurate and interpretable prediction of the properties of polycrystalline materials},
	volume = {7},
	issn = {2057-3960},
	doi = {10.1038/s41524-021-00574-w},
	language = {en},
	journal = {Computational Materials},
	author = {Dai, Minyi and Demirel, Mehmet F. and Liang, Yingyu and Hu, Jia-Mian},
	year = {2021},
	pages = {103},
}

@article{ge_numerical_2024,
	title = {Numerical {Assessment} of {Effective} {Elastic} {Properties} of {Needled} {Carbon}/{Carbon} {Composites} {Based} on a {Multiscale} {Method}},
	volume = {10},
	issn = {2311-5629},
	doi = {10.3390/c10030085},
	language = {en},
	journal = {Journal of Carbon Research},
	author = {Ge, Jian and Chao, Xujiang and Hu, Haoteng and Tian, Wenlong and Li, Weiqi and Qi, Lehua},
	year = {2024},
	pages = {85},
}

@article{yacouti_integrated_2025,
	title = {Integrated convolutional and graph neural networks for predicting mechanical fields in composite microstructures},
	volume = {190},
	issn = {1359835X},
	doi = {10.1016/j.compositesa.2024.108618},
	language = {en},
	journal = {Composites Part A: Applied Science and Manufacturing},
	author = {Yacouti, Marwa and Shakiba, Maryam},
	year = {2025},
	pages = {108618},
}

@article{vitulyova_hybrid_2025,
	title = {A {Hybrid} {Approach} {Using} {Graph} {Neural} {Networks} and {LSTM} for {Attack} {Vector} {Reconstruction}},
	volume = {14},
	issn = {2073-431X},
	doi = {10.3390/computers14080301},
	language = {en},
	journal = {Computers},
	author = {Vitulyova, Yelizaveta and Babenko, Tetiana and Kolesnikova, Kateryna and Kiktev, Nikolay and Abramkina, Olga},
	year = {2025},
	pages = {301},
}

@article{chinesta_short_2011,
	title = {A {Short} {Review} on {Model} {Order} {Reduction} {Based} on {Proper} {Generalized} {Decomposition}},
	volume = {18},
	issn = {1886-1784},
	doi = {10.1007/s11831-011-9064-7},
	journal = {Archives of Computational Methods in Engineering},
	author = {Chinesta, Francisco and Ladeveze, Pierre and Cueto, Elías},
	year = {2011},
	pages = {395--404},
}

@article{yu_3d_2020,
	title = {{3D} microstructural characterization and mechanical properties determination of short basalt fiber-reinforced polyamide 6,6 composites},
	volume = {187},
	issn = {13598368},
	doi = {10.1016/j.compositesb.2020.107839},
	language = {en},
	journal = {Composites Part B: Engineering},
	author = {Yu, Siwon and Hwang, Jun Yeon and Hong, Soon Hyung},
	year = {2020},
	pages = {107839},
}

@article{Humberto_2025,
title = {Microstructure and damage evolution in short carbon fibre 3D-printed composites during tensile straining},
journal = {Composites Part B: Engineering},
volume = {292},
pages = {112073},
year = {2025},
issn = {1359-8368},
doi = {10.1016/j.compositesb.2024.112073},
author = {José Humberto S. Almeida and Arttu Miettinen and Fabien Léonard and Brian G. Falzon and Philip J. Withers},
}

@article{arif_multiscale_2014,
	title = {Multiscale fatigue damage characterization in short glass fiber reinforced polyamide-66},
	volume = {61},
	issn = {13598368},
	doi = {10.1016/j.compositesb.2014.01.019},
	language = {en},
	journal = {Composites Part B: Engineering},
	author = {Arif, M.F. and Saintier, N. and Meraghni, F. and Fitoussi, J. and Chemisky, Y. and Robert, G.},
	year = {2014},
	pages = {55--65},
}

@article{gulmez_quantification_2023,
	title = {Quantification of structural response and edge orientation of {Chopped} {Tape} {Thermoplastic} {Composites} in net-shaped specimens},
	volume = {321},
	issn = {02638223},
	doi = {10.1016/j.compstruct.2023.117302},
	language = {en},
	journal = {Composite Structures},
	author = {Gulmez, Deniz Ezgi and Maldonado, Jesus and Masania, Kunal and Sinke, Jos and Dransfeld, Clemens},
	year = {2023},
	pages = {117302},
}

@article{Horst_1996,
author = {Horst, Jaap and Spoormaker, J.},
year = {1997},
pages = {3641-3651},
title = {Fatigue fracture mechanism and fractography of short-glassfibre-reinforced polyamide 6},
volume = {32},
journal = {Journal of Materials Science},
doi = {10.1023/A:1018634530869}
}

@article{summerscales_voronoi_2001,
	title = {Voronoi cells, fractal dimensions and fibre composites},
	volume = {201},
	issn = {0022-2720, 1365-2818},
	doi = {10.1046/j.1365-2818.2001.00841.x},
	language = {en},
	journal = {Journal of Microscopy},
	author = {Summerscales, J. and Guild, F. J. and Pearce, N. R. L. and Russell, P. M.},
	year = {2001},
	pages = {153--162},
}

@article{Senthilnathan_2024,
	title = {Comparison and validation of stochastic microstructure characterization and reconstruction: Machine learning vs. deep learning methodologies},
	volume = {278},
	issn = {1359--6454},
	doi = {10.1016/j.actamat.2024.120220},
	language = {en},
	journal = {Acta Materialia},
	author = {A. Senthilnathan and V. Saseendran and P. Acar and N. Yamamoto and V. Sundararaghavan},
	year = {2024},
	pages = {120220},
}

@article{billah_2024,
	title = {Experimental Investigation of Short Fiber Reinforced Thermoplastic Composite Joining in an Integrated Additive Manufacturing and Compression Molding System},
	volume = {1},
	doi = {10.1115/MSEC2024-125490},
	journal = {Additive Manufacturing; Advanced Materials Manufacturing; Biomanufacturing; Life Cycle Engineering },
	author = {Billah, Kazi Md Masum and Kumar, Vipin and Rathod, Neel and Phadatare, Akash and Nuttall, David and Smith, Tyler and Vaidya, Uday and Kim, Seokpum and Kunc, Vlastimil and Hassen, Ahmed Arabi},
	year = {2024},
	pages = {V001T02A013},
}

@article{Abhilash_2024,
author = {Nagaraja, Abhilash and BR, Abhiram and Ramesh, Chiranthan and Ghosh, Debraj and Gururaja, Suhasini},
title ={Multiscale modeling for mechanical property estimation of polymer matrix composites with nano- and micro-scale reinforcements},
journal = {Journal of Reinforced Plastics and Composites},
volume = {45},
pages = {1001-1021},
year = {2024},
doi = {10.1177/07316844241291718},}

@article{polym15010234,
author = {Bandinelli, Francesco and Peroni, Lorenzo and Morena, Alberto},
title ={Elasto-Plastic Mechanical Modeling of Fused Deposition 3D Printing Materials},
journal = {Polymers},
volume = {15},
pages = {234},
year = {2023},
doi = {10.3390/polym15010234},}

@article{Krairi2014,
  author  = {Krairi, A. and Doghri, I.},
  title   = {A thermodynamically-based constitutive model for 
             thermoplastic polymers coupling viscoelasticity, 
             viscoplasticity and ductile damage},
  journal = {International Journal of Plasticity},
  volume  = {60},
  pages   = {163--181},
  year    = {2014},
  doi     = {10.1016/j.ijplas.2014.04.010}
}

@article{Voce_1948,
author = {Voce E.},
title ={The Relationship between Stress and Strain for Homogeneous Deformation},
journal = {Journal of the Institute of Metals},
volume = {74},
pages   = {537--562},
year = {1948},
crid = {1570854176063010304},}

@article{WANG2013204,
author = {L. Wang and Z. Wang and S.M. Dong and W. Zhang and Y. Wang},
title ={Finite element simulation of stress distribution and development of Cf/SiC ceramic–matrix composite coated with single layer SiC coating during thermal shock},
journal = {Composites Part B: Engineering},
volume = {51},
pages = {204-214},
year = {2013},
doi = {10.1016/j.compositesb.2013.03.028},}

@article{Marshall_1985,
author = {D.B. Marshall and B.N. Cox and A.G. Evans},
title ={The Mechanics of Matrix Cracking in Brittle-Matrix Fiber Composites.},
journal = {Acta Metallurgica},
volume = {33},
pages = {2013-2021},
year = {1985},
doi = {10.1016/0001-6160(85)90124-5},}

@article{Marshall_1987,
title = {Tensile fracture of brittle matrix composites: Influence of fiber strength},
journal = {Acta Metallurgica},
volume = {35},
pages = {2607-2619},
year = {1987},
issn = {0001-6160},
doi = {10.1016/0001-6160(87)90260-4},
author = {D.B. Marshall and B.N. Cox},
}

@article{Leon_2009,
title = {Micromechanical modeling of damage and fracture of unidirectional fiber reinforced composites: A review},
journal = {Computational Materials Science},
volume = {44},
pages = {1351-1359},
year = {2009},
issn = {0927-0256},
doi = {10.1016/j.commatsci.2008.09.004},
author = {Leon Mishnaevsky and Povl Brøndsted},
}

@article{Lanids_1999,
author = {C. M. Landis and M. A. McGlockton and R. M. McMeeking},
title ={An Improved Shear Lag Model for Broken Fibers in Composite Materials},
journal = {Journal of Composite Materials},
volume = {33},
pages = {667-680},
year = {1999},
doi = {10.1177/002199839903300704},
}

@article{NAIRN199763,
title = {On the use of shear-lag methods for analysis of stress transfer in unidirectional composites},
journal = {Mechanics of Materials},
volume = {26},
pages = {63-80},
year = {1997},
issn = {0167-6636},
doi = {10.1016/S0167-6636(97)00023-9},
author = {John A. Nairn},
}

@article{Cox_1952,
doi = {10.1088/0508-3443/3/3/302},
year = {1952},
month = {mar},
volume = {3},
pages = {72},
author = {H L Cox},
title = {The elasticity and strength of paper and other fibrous materials},
journal = {British Journal of Applied Physics},
}

@article{Abhilash_2020,
author = {Nagaraja Abhilash and Gururaja Suhasini},
year = {2020},
title = {Effect of Chemical Vapor Infiltration Induced Matrix Porosity on the Mechanical Behavior of Ceramic Matrix Minicomposites},
volume = {6},
pages = {041005 },
journal = {ASCE-ASME J Risk and Uncert in Engrg Sys Part B Mech Engrg},
doi = {10.1115/1.4047465}
}

@article{Pathak_ffems_2026,
author = {Pathak, Pharindra and Bristy, Kaniz Fatema and Kumar, Vipin and Khonsari Michael M. and Gururaja, Suhasini},
title ={On the Influence of Cyclic Loading Frequency on Fatigue Limit of Short‐Fiber Thermoplastics},
journal = {Fatigue and Fracture of Engineering Materials and Structures},
volume = {45},
pages = {1001-1021},
year = {2026},
doi = {10.1111/ffe.70241},}

@article{kumar_high-performance_2021,
    title = {High-performance molded composites using additively manufactured preforms with controlled fiber and pore morphology},
    volume = {37},
    issn = {2214-8604},
    doi = {10.1016/j.addma.2020.101733},
	pages = {101733},
    journal = {Additive Manufacturing},
    author = {V. Kumar and S. P. Alwekar and V. Kunc and E. Cakmak and V. Kishore and T. Smith and J. Lindahl and U. Vaidya and C. Blue and M. Theodore and S. Kim and A. A. Hassen},
    year = {2021},
}

@article{pathak_porosity_2025,
    title = {Porosity evolution in additively manufactured compression molded short fiber thermoplastics under cyclic loading: Insights from micro-computed tomography and infrared thermography},
    volume = {194},
    issn = {1359-835X},
    doi = {10.1016/j.compositesa.2025.108881},
	pages = {108881},
    journal = {Composites Part A: Applied Science and Manufacturing},
    author = {P. Pathak and S. Gururaja},
    year = {2025},
}

@article{Almeida,
    title = {Effect of Void Content on the Strength of Composite Laminates},
    volume = {28},
    issn = {0263-8223},
    doi = {10.1016/0263-8223(94)90044-2},
	pages = {139-148},
    journal = {Composite Structures},
    author = {S. F. Almeida and Z. S. Neto},
    year = {1994},
}

@article{stamopoulos_evaluation_2016,
    title = {Evaluation of porosity effects on the mechanical properties of carbon fiber-reinforced plastic unidirectional laminates by X-ray computed tomography and mechanical testing},
    volume = {50},
    issn = {0021-9983},
    doi = {10.1177/0021998315602049},
	pages = {2087–98},
    journal = {Journal of Composite Materials},
    author = {A. Stamopoulos and K. Tserpes and P. Prucha and D. Vavrik},
    year = {2016},
}

@article{Pathak_Comp,
	title = {Examining infrared thermography based approaches to rapid fatigue characterization of additively manufactured compression molded short fiber thermoplastic composites},
	volume = {351},
	issn = {0263-8223},
	doi = {10.1016/j.compstruct.2024.118610},
	journal = {Composites Structure},
	author = {P. Pathak and S. Gururaja and V. Kumar and D. Nuttall and A. Mahmoudi and M.M. Khonsari and U. Vaidya},
	year = {2025},
    pages={118610},
}

@article{Mozaffar_2019,
author = {M. Mozaffar  and R. Bostanabad  and W. Chen  and K. Ehmann  and J. Cao  and M. A. Bessa },
title = {Deep learning predicts path-dependent plasticity},
journal = {Proceedings of the National Academy of Sciences},
volume = {116},
pages = {26414-26420},
year = {2019},
doi = {10.1073/pnas.1911815116},}

@article{Ghaboussi_1991,
author = {J. Ghaboussi  and J. H. Garrett  and X. Wu },
title = {Knowledge‐Based Modeling of Material Behavior with Neural Networks},
journal = {Journal of Engineering Mechanics},
volume = {117},
pages = {132-153},
year = {1991},
doi = {10.1061/(ASCE)0733-9399(1991)117:1(132)},}

@article{Burigede_2023,
title = {Learning macroscopic internal variables and history dependence from microscopic models},
journal = {Journal of the Mechanics and Physics of Solids},
volume = {178},
pages = {105329},
year = {2023},
issn = {0022-5096},
doi = {10.1016/j.jmps.2023.105329},
author = {Burigede Liu and Eric Ocegueda and Margaret Trautner and Andrew M. Stuart and Kaushik Bhattacharya},
}

@article{Head_2003,
  title = {Deformation of Cross-Linked Semiflexible Polymer Networks},
  author = {Head, David A. and Levine, Alex J. and MacKintosh, F. C.},
  journal = {Phys. Rev. Lett.},
  volume = {91},
  issue = {10},
  pages = {108102},
  year = {2003},
  publisher = {American Physical Society},
  doi = {10.1103/PhysRevLett.91.108102},
}

@book{torquato_random_2002,
	title = {Random {Heterogeneous} {Materials}: {Microstructure} and {Macroscopic} {Properties}},
	volume = {55},
	doi = {10.1115/1.1483342},
	author = {Torquato, Salvatore},
	year = {2002},
	note = {Publication Title: Applied Mechanics Reviews - APPL MECH REV},
    publisher= {Springer Nature}
}

@misc{AdamW,
      title={Decoupled Weight Decay Regularization}, 
      author={Ilya Loshchilov and Frank Hutter},
      year={2019},
      eprint={1711.05101},
      archivePrefix={arXiv},
      primaryClass={cs.LG},
}

@book{Bengio2012,
	title = {Practical Recommendations for Gradient-Based Training of Deep Architectures},
    booktitle = {Neural Networks: Tricks of the Trade: Second Edition},
    publisher={Springer Berlin Heidelberg},
	volume = {7700},
	doi = {10.1007/978-3-642-35289-8_26},
	author = {Bengio, Yoshua},
	year = {2012},
	pages = {437--478},
}

@inproceedings{kumar2022anisotropic,
author = {Pokkalla, Deepak Kumar and Kumar, Vipin and Jo, Eonyeon and Hassen, Ahmed Arabi and Cakmak, Ercan and Alwekar, Shailesh and Kunc, Vlastimil and Vaidya, Uday and Baid, Harsh and Kim, Seokpum},
year = {2022},
pages = {46},
booktitle={Nondestructive Characterization and Monitoring of Advanced Materials, Aerospace, Civil Infrastructure, and Transportation XVI},
publisher={SPIE},
title = {Anisotropic mechanical properties of polymer composites from a hybrid additive manufacturing-compression molding process using x-ray computer tomography},
doi = {10.1117/12.2614620}
}

@article{liu2022effect,
author = {Liu, Zhu and Lei, Yongpeng and Zhang, Xiangyang and Kang, Zhenhang and Zhang, Jifeng},
title = {Effect Mechanism and Simulation of Voids on Hygrothermal Performances of Composites},
journal = {Polymers},
volume = {14},
year = {2022},
issn = {2073-4360},
doi = {10.3390/polym14050901}
}
\end{singlespace}
\newpage

\appsec{Parametric Finite-Element Validation of the Edge Weight Function}
\label{app:A1}
The edge weight function (Equation~\ref{edge_weight}) embeds two physically motivated assumptions regarding pairwise fiber--fiber load transfer: (i) matrix-mediated stress transfer between proximal fiber ends decays with increasing fiber--fiber separation, and (ii) load-transfer efficiency depends on relative fiber orientation through a dominant cosine-dependent axial shear-lag coupling and a residual orientation-independent transverse stress concentration. To confirm the physical basis of both terms in the CF-ABS system at geometries consistent with $\mu$-CT observations, a systematic parametric finite-element study was conducted using a two-fiber RUC model. Because Equation~\ref{edge_weight} governs physics at the pairwise level (multi-fiber effects are handled through GNN message-passing aggregation), the two-fiber RUC represents the minimal and most direct unit for validating its functional form.

\begin{figure}[htp!]
    \centering
    \includegraphics[width=1\linewidth]{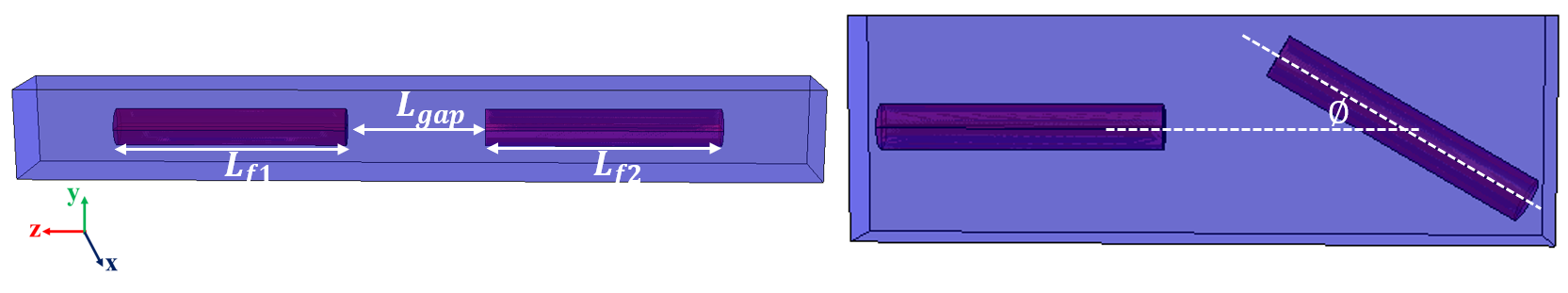}
    \caption{Two-fiber RUC model of CF-ABS used for parametric shear-lag studies. \textbf{Left}: $\phi = 0^\circ$ with fiber lengths $L_{f1}, L_{f2}$ and end-to-end separation $L_{\text{gap}}$ labeled. \textbf{Right}: $\phi = 30^\circ$ case.}
    \label{fig:A2_1}
\end{figure}

\begin{figure}[htp!]
    \centering
    \includegraphics[width=1\linewidth]{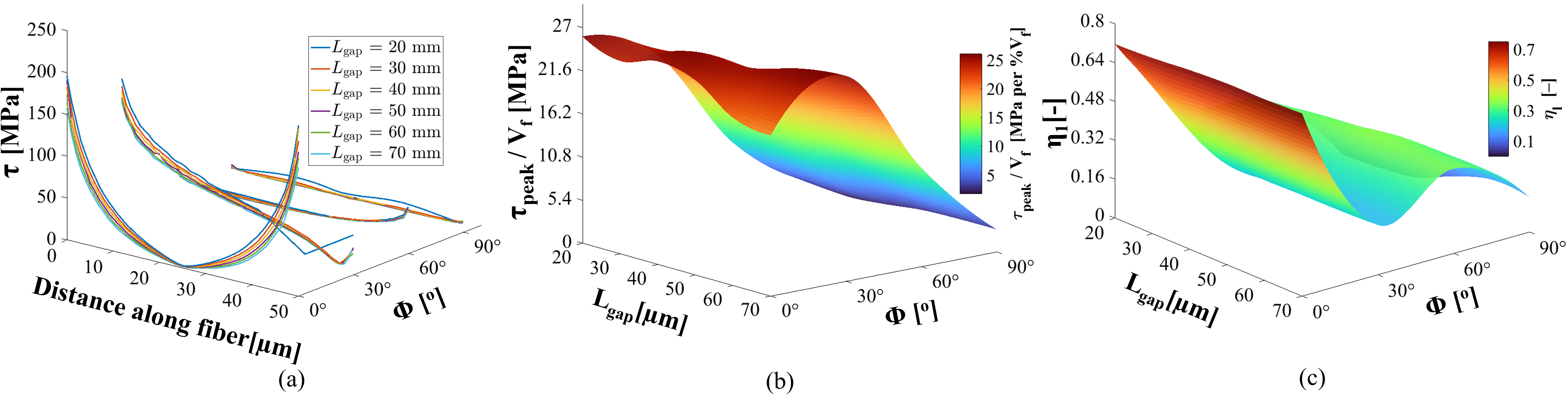}
    \caption{Shear-lag response of two-fiber CF-ABS RVEs across all 24 parametric cases (four orientations $\times$ six gap distances).
    (a)~Absolute interfacial shear stress $\tau(z)$ profiles extracted along the matrix--fiber interface path. Each set of curves corresponds to one fiber orientation ($\phi = 0^{\circ}$, $30^{\circ}$, $60^{\circ}$, $90^{\circ}$), with individual curves representing gap distances $L_{\mathrm{gap}} = 20$--$70$~mm. %
    (b)~$V_f$-normalised peak interfacial shear stress $\tau_{\mathrm{peak}}/V_f$ [MPa per $\%V_f$] as a three-dimensional surface over the $L_{\mathrm{gap}}$--$\phi$ parameter space. Normalization by local fiber volume fraction removes the geometric $V_f$ variation introduced by the tilted-fiber box geometry.
    (c)~ Fiber length efficiency $\eta_l$ estimated from the data-driven exponential decay of the normalised $\tau$ profile, shown as a three-dimensional surface over the $L_{\mathrm{gap}}$--$\phi$ parameter space.}
    \label{fig:A2_2}
\end{figure}

\paragraph{Two-fiber RUC model geometry}
The model consists of two cylindrical carbon fibers of equal length $L_{f1} = L_{f2} = 50~\mu$m and radius $r_f = 4~\mu$m embedded in an ABS matrix cuboid (Figure~\ref{fig:A2_1}). The first fiber is fixed in all analyses; the second is varied such that the end-to-end separation $L_{gap}$ ranges from 20 to 70~$\mu$m in six increments, and the relative orientation angle $\phi$ takes values of $0^\circ$, $30^\circ$, $60^\circ$, and $90^\circ$, yielding 24 parametric cases. All cases were analyzed under uniaxial tensile strain $\varepsilon_0 = 4\%$ applied along the primary fiber longitudinal direction, consistent with the homogenization protocol of Section~2.3.3. For angled configurations, RUC dimensions were adjusted to accommodate the tilted geometry while keeping fiber dimensions fixed, resulting in orientation-dependent fiber volume fractions: $V_f = 7.86\%$ at $0^\circ$, $6.60\%$ at $30^\circ$, $5.60\%$ at $60^\circ$, and $4.83\%$ at $90^\circ$.
\paragraph{Interfacial shear stress extraction}
For each parametric case, the interfacial shear stress resultant:
\begin{equation}
  \tau
  =
  \sqrt{\tau_{xy}^2 + \tau_{yz}^2 + \tau_{zx}^2}
  \label{eq:tau_resultant}
\end{equation}
was extracted at the first matrix integration point immediately adjacent to the fiber--matrix interface along the full fiber axis from $z=0$ to $z=L_f$. The peak interfacial shear stress $\tau_{peak}$ was identified and normalized by local $V_f$ to enable consistent comparison across orientations (Figure~\ref{fig:A2_2}b).

\paragraph{Validation of the distance-decay term}
Figure~\ref{fig:A2_2}a presents $\tau$ profiles for all 24 cases. Across all four orientations, interfacial shear concentrated at both fiber termini, confirming that end-to-end proximity governs pairwise load transfer regardless of orientation. At $0^\circ$ and $30^\circ$, $\tau_{peak}$ decreased monotonically with increasing $L_{gap}$ (15--22\% reduction at $0^\circ$ between 20 and 70~$\mu$m), directly confirming the exponential decay term $\exp(-d_{pq}/L_c)$ in Equation~ \ref{edge_weight}. At $60^\circ$ and $90^\circ$, the gap effect was negligible (less than 8\%), confirming that the distance-decay mechanism is physically operative only when fibers participate in axial shear-lag-governed load transfer---precisely the condition captured by the orientation-dependent component.

\paragraph{Validation of the two-component orientation factor}
Figure~\ref{fig:A2_2}b reveals two physically distinct orientation regimes motivating the $(0.9|\mathbf{t}_p \cdot \mathbf{t}_q| + 0.10)$ structure in Equation~9. The $V_f$-normalized peak stress decreased systematically with misalignment, from 25--32~MPa per \%$V_f$ at $0^\circ$ to 2--7~MPa per \%$V_f$ at $90^\circ$---an 8--10-fold reduction that validates the dominant cosine-dependent axial shear-lag term. Critically, even at $90^\circ$ a non-negligible residual of 2--7~MPa per \%$V_f$ persisted, arising from transverse stress concentration due to fiber--matrix stiffness mismatch rather than axial shear-lag. This residual was gap-insensitive (less than 8\% variation across the full $L_{gap}$ range), confirming its orientation-independent character. The observed residual fraction of 0.08--0.22 relative to the $0^\circ$ maximum directly yields the constant 0.10 in Equation~\ref{edge_weight}, with the complementary 0.9 assigned to the cosine-dependent term.

\paragraph{Fiber length efficiency}
The fiber length efficiency $\eta_l$, presented as a three-dimensional surface in Figure~\ref{fig:A2_2}c, quantifies the fraction of theoretical fiber stiffness contribution activated over the actual fiber length as directly observed from the two-fiber FEA results. Values of $\eta_l = 0.72$--$0.75$ for $0^\circ$ fiber pairs confirm that, at the fiber lengths and separations present in the $\mu$-CT microstructure, pairwise fiber--fiber interaction is strongly active and a large fraction of available reinforcement efficiency is realized through end-to-end load transfer. For $30^\circ$ fibers, $\eta_l = 0.20$--$0.38$, consistent with the intermediate-orientation behavior described above.

\appsec{GNN-LSTM Training Configuration}
\label{app:A2}

Table~\ref{tab:A2} consolidates the hyperparameters used for GNN-LSTM optimization (described in Section~\ref{GNN_LSTM}).

\begin{table}[h]
    \centering
    \caption{Training hyperparameters used for GNN--LSTM model optimization.}
    \label{tab:A2}
    \begin{tabular}{ll}
        \hline
        \textbf{Parameter} & \textbf{Value} \\
        \hline
        Optimizer & AdamW \\
        Initial learning rate & $5 \times 10^{-4}$ \\
        Weight decay & $2 \times 10^{-4}$ \\
        LR scheduler & ReduceLROnPlateau (factor 0.5, patience 15 epochs) \\
        Minimum learning rate & $10^{-6}$ \\
        Gradient clip norm & 1.0 \\
        Batch size & 1 \\
        Maximum epochs & 800 \\
        Early stopping patience & 57 epochs (validation loss) \\
        Primary loss & Strain-weighted Huber ($\delta = 0.1$) \\
        $\lambda_{\text{mono}}$ & 0.05 \\
        $\lambda_{\text{smooth}}$ & 0.001 \\
        $\lambda_{\text{zero}}$ & 0.02 \\
        $\lambda_{\text{aux}}$ & 0.3 \\
        $\lambda_{\text{peak}}$ & 0.5 \\
        Train / Val / Test split & 70\% / 15\% / 15\% (stratified random shuffle, seed 42) \\
        GNN hidden / embedding dimension & 64 / 16 \\
        GNN layers & 2 (GCNConv + residual + LN + Dropout 0.2) \\
        LSTM hidden dimension / layers & 128 / 2 (inter-layer Dropout 0.25) \\
        Auxiliary head & Linear(16,32), ReLU, Dropout(0.1), Linear(32,4) \\
        RF estimators & 300 (\texttt{MultiOutputRegressor}) \\
        \hline
    \end{tabular}
\end{table}

Two implementation choices warrant brief justification. A batch size of 1 was used because Voronoi cells contain variable numbers of fibers, which would otherwise require variable-size padding for batched graph processing; sequential processing avoids this overhead while gradient norm clipping maintains stability. The 500-cell dataset was partitioned using stratified random shuffling rather than a sequential split, since a sorted split would risk grouping all low-fiber-count or high-$V_f$ cells into a single subset, producing an out-of-distribution validation set and overoptimistic generalization estimates.

\appsec{Full Six-Mode Constitutive Response of the Discrete Cell}
\label{app:A3}

\begin{figure}[htp!]
    \centering
    \includegraphics[width=1\linewidth]{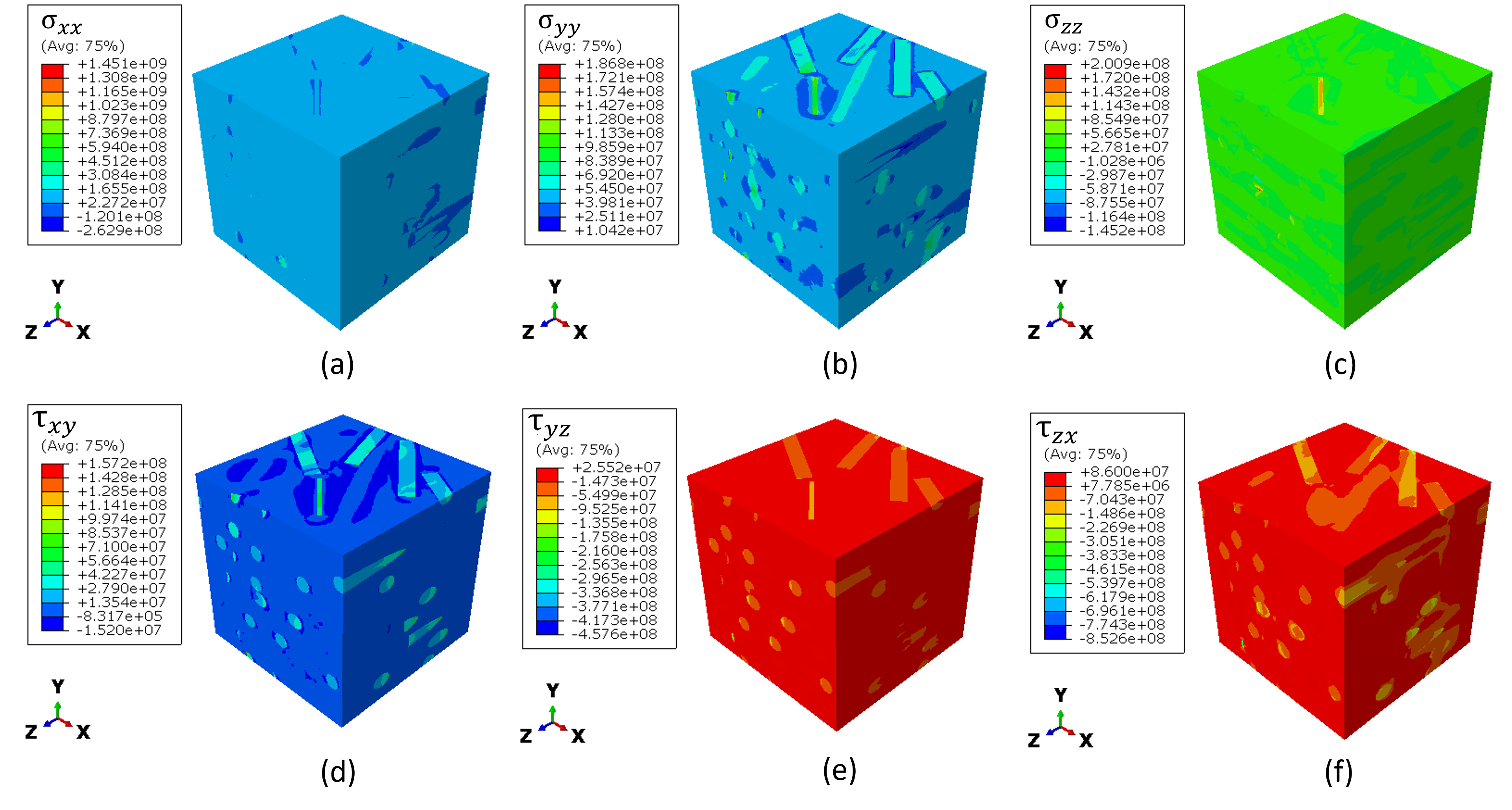}
    \caption{Full six-mode constitutive response of a representative discrete Voronoi cell from finite-element homogenization. Von Mises stress fields are shown under: (a) normal loading in the $x$-direction ($\sigma_{xx}$); (b) transverse loading in the $y$-direction ($\sigma_{yy}$); (c) transverse loading in the $z$-direction ($\sigma_{33}$); (d) axial shear in the $x$--$y$ plane ($\tau_{12}$); (e) axial shear in the $y$--$z$ plane ($\tau_{23}$); (f) axial shear in the $x$--$z$ plane ($\tau_{13}$). The contrasting stress patterns under normal versus shear loading confirm that the cell’s constitutive response is strongly anisotropic and topology-sensitive.}
    \label{fig:A4_1}
\end{figure}

To verify that the discrete-cell FE homogenization produces physically consistent constitutive responses under all six independent loading modes, Figure \ref{fig:A4_1} presents the full von Mises stress fields for a representative Voronoi cell subjected to three normal and three shear loading cases. Under normal loading in the $y$- and $z$-directions (Figure~\ref{fig:A4_1}(a–b), stress is distributed relatively uniformly through the matrix, consistent with matrix-dominated transverse response and the comparatively small contribution of off-axis fibers to transverse stiffness. Under normal loading in the $x$-direction (fiber-dominated, Figure~\ref{fig:A4_1}c), pronounced stress localization occurs along fiber axes, confirming efficient axial load transfer.

Under shear loading (Figure~\ref{fig:A4_1}(d–f)), stress concentrations developed in matrix ligaments between adjacent fibers, consistent with shear-lag-governed inter-fiber coupling. These include one transverse shear configuration acting within the cross-sectional $z$--$y$ plane, and two axial shear configurations acting across the longitudinal $y$--$z$ and $x$--$z$ fiber planes. These results confirm that the discrete-cell FE implementation correctly captures the anisotropic, microstructure-sensitive constitutive behavior that the GNN-LSTM surrogate is trained to reproduce.

\newpage
\supsec{Sphere-Sliding Visualization}
\label{sup:A}
Please find the following supplemental material available below:
\href{https://tigermailauburn-my.sharepoint.com/:v:/g/personal/pzp0057_auburn_edu/IQDk6mAbJ_78SqZB0yzK5vwJAcAzcsbCv30MR8ja6wenuBg?nav=eyJyZWZlcnJhbEluZm8iOnsicmVmZXJyYWxBcHAiOiJPbmVEcml2ZUZvckJ1c2luZXNzIiwicmVmZXJyYWxBcHBQbGF0Zm9ybSI6IldlYiIsInJlZmVycmFsTW9kZSI6InZpZXciLCJyZWZlcnJhbFZpZXciOiJNeUZpbGVzTGlua0NvcHkifX0&e=2K3hAY}{Sphere-sliding video showing how the fibers number and orientation has been defined for each discrete model}
 \end{doublespace} 
\end{document}